\documentclass[acmtog]{acmart}

\usepackage[ruled,vlined,linesnumbered]{algorithm2e} %
\usepackage{multirow} 
\usepackage[inline]{enumitem}
\usepackage{verbatim}

\SetAlFnt{\small}
\SetAlCapFnt{\small}
\SetAlCapNameFnt{\small}
\SetAlCapHSkip{0pt}

\copyrightyear{2026}
\acmYear{2026}
\setcopyright{cc}
\setcctype{by}
\acmConference[SIGGRAPH Conference Papers '26]{Special Interest Group on Computer Graphics and Interactive Techniques Conference Conference Papers}{July 19--23, 2026}{Los Angeles, CA, USA}
\acmBooktitle{Special Interest Group on Computer Graphics and Interactive Techniques Conference Conference Papers (SIGGRAPH Conference Papers '26), July 19--23, 2026, Los Angeles, CA, USA}
\acmDOI{10.1145/3799902.3811093}
\acmISBN{979-8-4007-2554-8/2026/07}

\usepackage[table]{xcolor}

\usepackage{xspace}
\SetAlFnt{\small}
\SetAlCapFnt{\small}
\SetAlCapNameFnt{\small}
\SetAlCapHSkip{0pt}

\newcommand{\method}{\textsc{Go-with-the-Track}\xspace}

\newcommand{\eg}{e.g.,}

\usepackage{cleveref}

\begin{document}
\title{Go-with-the-Track: Video Compositing and Motion Control with Point Tracking} 

\author{Koichi Namekata}
\orcid{0009-0001-6146-5659}
\affiliation{
    \institution{Netflix}
    \country{USA}
}
\affiliation{
    \institution{Eyeline Labs}
    \country{USA}
}
\affiliation{
    \institution{University of Oxford}
    \city{Oxford}
    \country{United Kingdom}
}
\email{koichi.namekata@eng.ox.ac.uk}

\author{Yash Kant}
\orcid{0009-0002-8347-4895}
\authornote{Equal Supervision.\\Koichi Namekata, Zhizheng Liu, Ryan Burgert, and Kuan Heng Lin performed this work during an internship at Netflix and Eyeline Labs.}
\affiliation{
    \institution{Netflix}
    \country{USA}
}
\affiliation{
    \institution{Eyeline Labs}
    \city{Los Angeles}
    \country{USA}
}
\email{ysh.kant@gmail.com}

\author{Zhizheng Liu}
\orcid{0009-0006-9426-3718}
\affiliation{
    \institution{Eyeline Labs}
    \country{USA}
}
\affiliation{
    \institution{University of California, Los Angeles}
    \city{Los Angeles}
    \country{USA}
}
\email{zhizheng@cs.ucla.edu}

\author{Ryan Burgert}
\orcid{0009-0008-5947-2076}
\affiliation{
    \institution{Netflix}
    \country{USA}
}
\affiliation{
    \institution{Eyeline Labs}
    \country{USA}
}
\affiliation{
    \institution{Stony Brook University}
    \country{USA}
}
\email{ryancentralorg@gmail.com}

\author{Yuancheng Xu}
\orcid{0000-0002-2254-5752}
\affiliation{
    \institution{Netflix}
    \country{USA}
}
\affiliation{
    \institution{Eyeline Labs}
    \city{Los Angeles}
    \country{USA}
}
\email{xuyuancheng0@gmail.com}

\author{Kuan Heng Lin}
\orcid{0009-0003-0605-5862}
\affiliation{
    \institution{Eyeline Labs}
    \country{USA}
}
\affiliation{
    \institution{Columbia University}
    \country{USA}
}
\email{jordan@cs.columbia.edu}

\author{Emmett Steven}
\orcid{0009-0004-2852-4444}
\affiliation{
    \institution{Netflix}
    \country{USA}
}
\email{esteven@netflix.com}

\author{Julien Philip}
\orcid{0000-0003-3125-1614}
\affiliation{
    \institution{Eyeline Labs}
    \country{United Kingdom}
}
\email{julienov.philip@gmail.com}

\author{Li Ma}
\orcid{0000-0002-6992-0089}
\affiliation{
    \institution{Eyeline Labs}
    \city{Los Angeles}
    \country{USA}
}
\email{lmaag@connect.ust.hk}

\author{Andrea Vedaldi}
\orcid{0000-0003-1374-2858}
\affiliation{
    \institution{University of Oxford}
    \city{Oxford}
    \country{United Kingdom}
}
\email{vedaldi@robots.ox.ac.uk}

\author{Paul Debevec}
\orcid{0000-0001-7381-2323}
\affiliation{
    \institution{Netflix}
    \city{Los Angeles}
    \country{USA}
}
\affiliation{
    \institution{Eyeline Labs}
    \city{Los Angeles}
    \country{USA}
}
\email{debevec@gmail.com}

\author{Ning Yu}
\authornotemark[1]
\orcid{0009-0004-6865-1325}
\affiliation{
    \institution{Netflix}
    \city{Los Angeles}
    \country{USA}
}
\affiliation{
    \institution{Eyeline Labs}
    \city{Los Angeles}
    \country{USA}
}
\email{ningyu.hust@gmail.com}

\makeatletter
\let\@authorsaddresses\@empty
\makeatother
\renewcommand{\shortauthors}{Namekata et al.}

\begin{abstract}

Filmmaking demands precise motion control and reference image compositing — capabilities that existing methods treat separately. Point-track-conditioned image-to-video models restrict content insertion to the first frame, while reference-to-video models lack fine-grained spatial-temporal control over how reference content integrates across frames.

We present \method, which unifies both capabilities by jointly conditioning on multiple reference images and \textit{reference-anchored point-tracks} — extending conventional point-tracks to explicitly establish correspondences between generated frames and reference images, thus enabling precise compositing and motion control throughout the video.

To achieve this, we introduce \textit{spatially-aware} point-track embeddings that encode the full sequence of point-track coordinates using a coordinate-wise MLP followed by temporal pooling. 
This representation captures the spatial characteristics of each point-track (serving as a unique identifier), while the embedding similarity correlates directly with spatial proximity, enhancing the model's ability to distinguish and associate point-tracks.
We inject these point-track embeddings into a video diffusion transformer via a lightweight adapter, resolving the pixel-to-patch resolution mismatch while avoiding the substantial motion detail loss inherent in naive point-track subsampling.

We use a hybrid training strategy to train jointly on dynamic, static, and synthetic scene video datasets to boost motion controllability. Experiments demonstrate that \method achieves superior motion and reference control in a single model and enables new capabilities: multi-reference conditioned video generation with point-track driven compositing, as well as camera control for both static and dynamic scenes.
Project Page: \url{https://eyeline-labs.github.io/Go-with-the-Track/}

\end{abstract}



\maketitle

\begin{figure*}[t!]
  \centering
  \includegraphics[width=\textwidth]{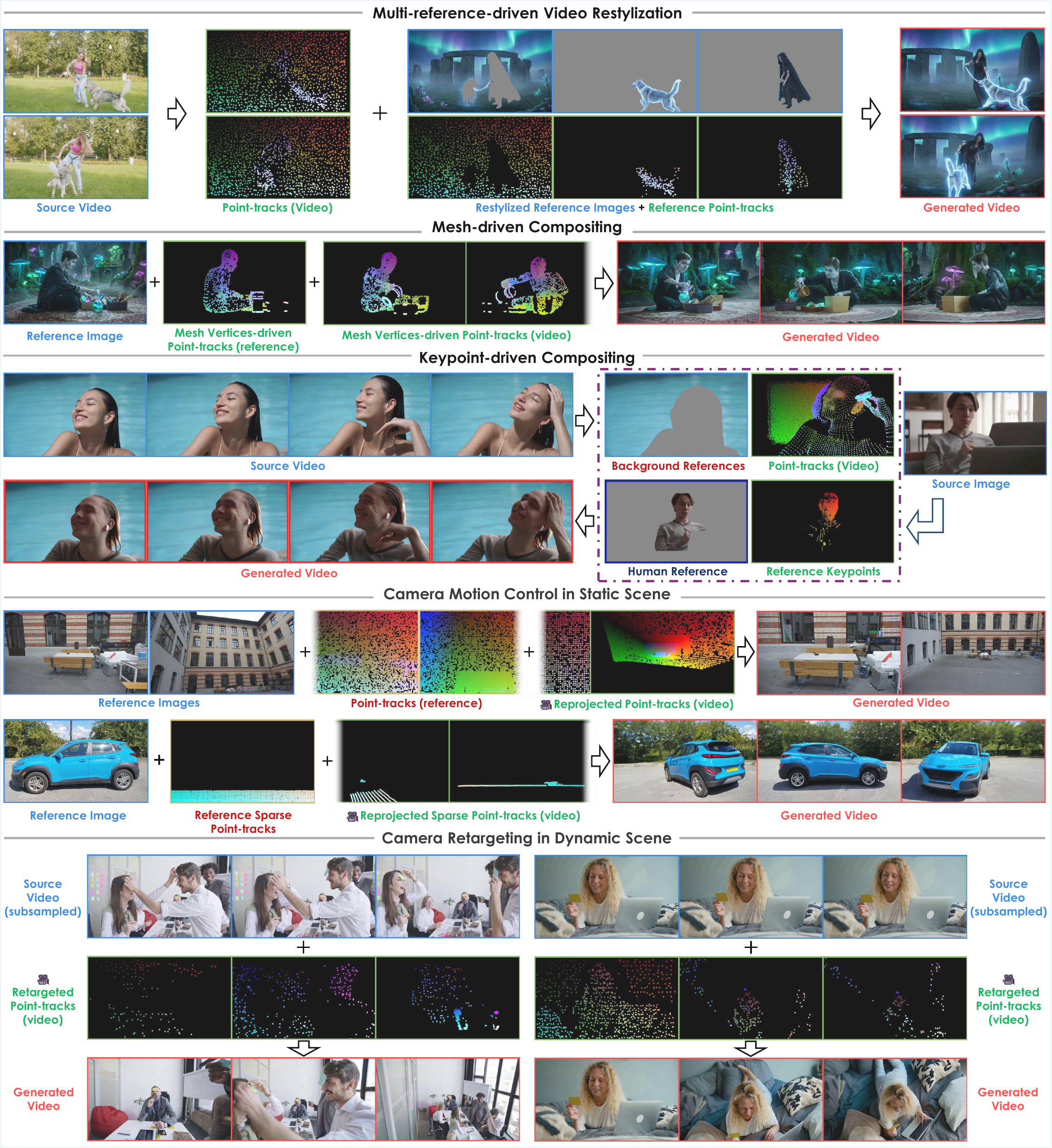}
  \caption{\textbf{Applications of \method.}  We present a simple and flexible video generation framework conditioned on multiple reference images and point-tracks that anchor both the generated frames and the references. Our approach enables: (1) motion-preserved video restylization using multiple reference images with point-tracks from off-the-shelf trackers; (2) we also support mesh- or keypoint-driven compositing; and (3) camera motion control  for both static and dynamic scenes with one or more reference images. Please refer to the supplementary webpage and the appendix figures  for additional results.}
  \label{fig:teaser}
\end{figure*}

\section{Introduction}
\label{sec:intro}
Recent progress in large-scale video generative foundation models~\cite{wan2025wan,kong2025hunyuanvideosystematicframeworklarge,genmo2024mochi,ray3_luma_ai,nvidia2025worldsimulationvideofoundation} has pushed video synthesis toward cinematic quality, enabling high-fidelity appearance, scene coherence, and dynamic motion rendering.
However, filmmaking is fundamentally about weaving characters and scenes through time and space; therefore, video creators increasingly need precise control over \textit{what appears in a scene} and \textit{how it moves} --- requirements that text prompts alone cannot satisfy. 
In creative workflows, visual appearance must remain consistent with reference materials (e.g., costumes, props), while the spatiotemporal development of the scene must be directed with intention and continuity (e.g., character motion, camera movement).

A growing body of point-track-conditioned video generative models~\cite{gu2025das,geng2025motionprompting,li2025trackdiffusion,ma2023trailblazer,qiu2024freetraj,lei2025ditraj,wang2023motionctrl,jin2025flovd,ji2025posetraj,hou2024trainingfreecamera,namekata2024sg, chu2025wan} use point-tracks that are anchored in the first frame. 
While this approach offers intuitive control over how the first frame evolves, it also implicitly assumes that \textit{all controllable elements are present in the initial frame} of the clip. 
As a consequence, these approaches struggle to compose elements (e.g. characters, props, or backgrounds) that enter or exit the scene naturally at different frames. 
In parallel, multi-reference and personalized video generative models~\cite{jiang2025vace,jiang2023videobooth,chen2023videodreamer,deng2025magref,sang2025lynx,wang2024animatelcm,chefer2024stillmoving,blattmann2023videoldm,chen2025multisubject,bian2025videoasprompt,liu2025kaleido} inject appearances from one or more images or videos. However, these approaches typically lack an explicit, editable representation of motion, which makes it difficult to control multiple reference elements in space and time. 

Filmmaking requires both appearance and motion controls to operate in a unified fashion, because characters, props, and environments must retain a consistent visual identity while entering, moving, interacting, and exiting at precisely chosen locations and times within the shot.
\textit{Towards this goal, we introduce \method, a model that enables joint control over multiple references as well as their spatiotemporal placement in the video.}  
Our key idea is to use point-tracks as spatiotemporal anchors that unify motion control and reference image insertion (compositing) within a single framework. 
While conventional point-tracks are defined merely as the 2D flow of a point strictly within the video frames, we extend this definition to \textbf{multi-reference anchored} point-tracks by explicitly linking each point-track to corresponding coordinates on the multiple reference images given as inputs. 
By establishing these point correspondences between the generated video frames and the reference images, each track instructs not only how an element moves across generated frames, but exactly what element of the reference image is anchored to that movement. 

A core challenge in achieving such multi-reference anchored point-track controllability is ensuring the model can distinctively associate a large number of conditioning point-tracks (e.g., up to $15,000$) between generated frames and multiple reference images, where they are not spatiotemporally continuous. 
Previous point-track-conditioned approaches~\cite{geng2025motionprompting, shin2025motionstream, burgert2025motionv2veditingmotionvideo} typically represent point-track identities using unique random embeddings; however, we find such spatially uncorrelated identifiers are suboptimal for precise  reference insertion.
We attribute this limitation to the inherent spatiotemporal discontinuity between generated frames and the reference images, where models cannot exploit spatial proximity (i.e., the smooth temporal evolution of positional coordinates within each point-track) to naturally link points belonging to the same point-track across frames.
As a result, the model is forced to match points solely by distinguishing dense random embeddings. 
This approach suffers from slow convergence and poor generalization as the number of conditioning point-tracks increases.

\begin{figure*}[t!]
  \centering
  \includegraphics[width=\textwidth]{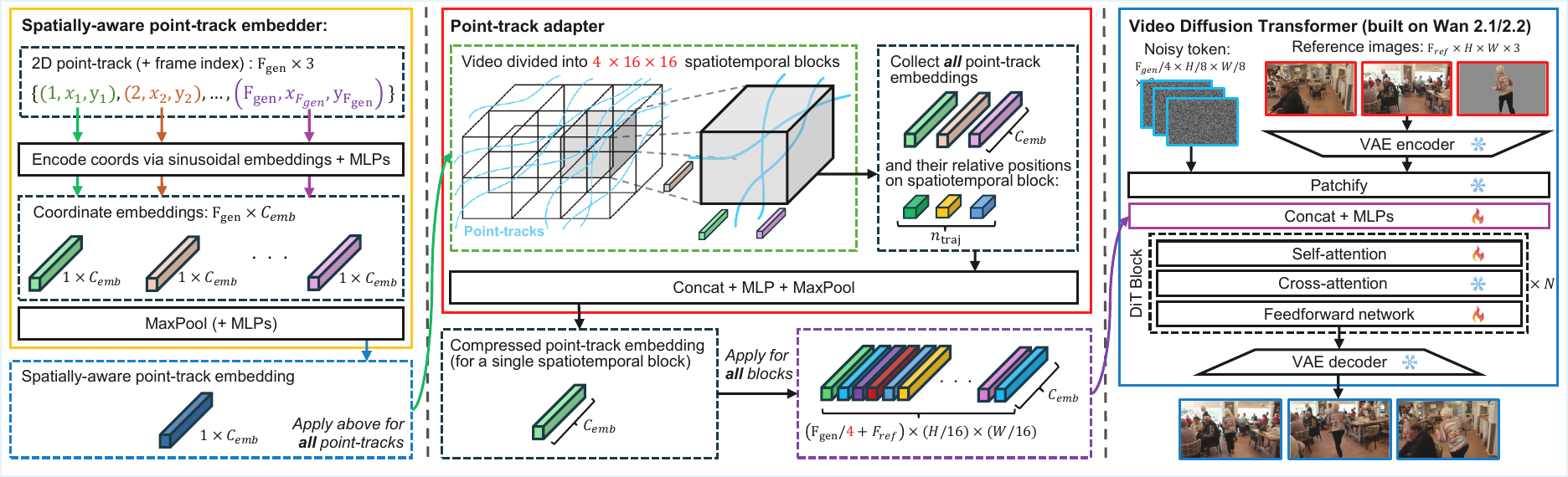} 

  \caption{\textbf{Design details of Go-with-the-Track.} (Left): \textit{Point-Track Embedder}: For each point-track, we encode its coordinate sequence using a shared coordinate-wise MLP, followed by temporal max-pooling. This produces a single spatially-aware point-track embedding that is virtually distributed across frames.  (Middle) \textit{Point-Track Adapter}: To downsample pixel-space point-track embeddings into a compressed patchified latent space, our lightweight adapter aggregates embeddings that fall within each $4 \times 16 \times 16$ spatiotemporal block. These embeddings are concatenated with their relative position and max pooled to obtain a single compressed point-track embedding per block. (Right) \textit{Video Diffusion Transformer:} The compressed point-track embeddings are channel-wise concatenated with the noisy latent video and VAE-encoded reference images. The combined representation is then processed by a partially fine-tuned video diffusion transformer to generate the final video.}
\label{fig:trajectory_architecture}

\end{figure*}

To address this issue, we devise a methodology to build \textbf{spatially-aware point-track embeddings}. 
Specifically, as illustrated in the left part of ~\cref{fig:trajectory_architecture}, we encode the point-track coordinates corresponding to generated video frames (where motion continuity is preserved) using a coordinate-wise MLP followed by temporal pooling. 
This formulation produces an embedding vector that serves as a unique identifier for each point-track while inherently representing its spatial characteristics. 
Crucially, this ensures that embedding similarity correlates directly with spatial proximity, providing the spatial cues, which facilitate the model to effectively associate points between the reference and generated frames, even when their raw coordinates are disjoint. 

Building upon the open-sourced video latent diffusion transformer~\cite{wan2025wan}, we introduce a \textbf{lightweight point-track adapter} that injects these point-track embeddings into the diffusion models.
The latent diffusion model operates in a highly compressed patchified latent space (downsampled $16\times16$ spatially and $4\times$ temporally), whereas our point-tracks are defined in the high-resolution pixel space.
To bridge this gap, we train a lightweight adapter that efficiently aggregates all point-tracks falling within each corresponding $4 \times 16 \times 16$ spatiotemporal block into a single compressed conditioning vector (the middle part of \cref{fig:trajectory_architecture}).
Crucially, we carefully design the adapter to avoid the significant loss of motion details inherent in the previous naive point-track subsampling methods~\cite{wang2025, chu2025wan, burgert2025, shin2025motionstream}, while maintaining minimal computational overhead.

\begin{figure*}[t!]
  \centering
  \includegraphics[width=0.9\textwidth]{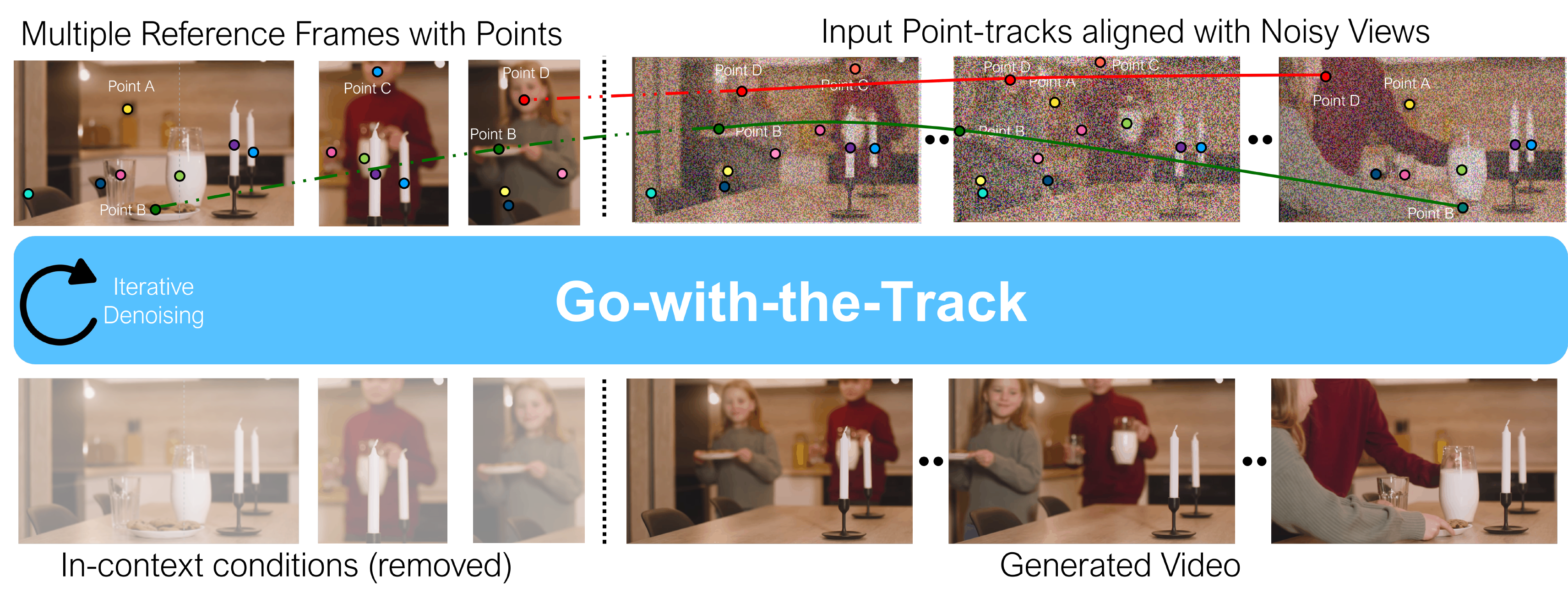}
  \caption{\textbf{\method overview.} Top: \method conditions on multiple reference images and point-track annotations spatially aligned with both reference and generated frames. 
Bottom: The resulting video after denoising.}
  \label{fig:pipeline}
\end{figure*}

Another challenge in training is the lack of video datasets annotated with accurate ground-truth point-tracks. 
Prior works~\cite{geng2025motionprompting,gu2025das,burgert2025motionv2veditingmotionvideo,chu2025wan,shin2025motionstream} rely on off-the-shelf point trackers~\cite{SpatialTracker,doersch2024} to create training data. However, this produces noisy point-tracks, weakening the model's ability to precisely follow point-track conditioning.
To address this issue, we utilize a mixture of synthetic datasets (i.e., PointOdyssey~\cite{zheng2023point}), static-scene datasets (i.e., DL3DV~\cite{ling2024dl3dv}, TartanAir~\cite{tartanair2020iros}), and high-quality dynamic-scene video datasets (e.g., OpenVidHD~\cite{nan2024openvid}, MiraData~\cite{ju2024miradatalargescalevideodataset}, OpenHumanVid~\cite{li2024openhumanvid}). 
Synthetic and static-scene datasets provide ground-truth point-tracks derived either from mesh vertices or from ground-truth depth and camera annotations, while dynamic scene datasets provide diverse and photorealistic video priors. These enable \method to acquire precise motion controllability under complex object and camera motions. 

In summary, our contributions include:
\begin{itemize}
    \item \textbf{Unified Motion Control and Compositing:} We identify that unification of motion control and multi-reference compositing unlocks powerful controllability in video generation, addressing practical filmmaking demands. By proposing reference-anchored point-tracks as a new conditioning paradigm, \method enables spatiotemporally controllable video generation from multiple reference images.
    \item {\textbf{Spatially-Aware Point-track Embeddings \& Adapter:} 
    We identify that random embeddings used in previous work are suboptimal for representing reference-anchored point-track conditioning and address this limitation with spatially-aware point-track embeddings. Coupled with a carefully designed lightweight adapter, our method effectively maps pixel-space point-track conditioning into compressed patchified latent space without substantial loss of motion details.}
    
    \item {\textbf{Hybrid Training Data:} We combine synthetic and static scene datasets on top of real video datasets. This hybrid training strategy reduces the reliance on noisy and error-prone point trackers, enabling the model to learn precise motion control while preserving strong photorealistic video priors.}
    \item \textbf{Wide Range of Applications:} {\method outperforms baselines on quantitative metrics and user studies, and unlocks a wide range of applications by leveraging the flexibility of our reference and point-track conditioning:}
    \begin{enumerate*}[label=(\arabic*)]
    \item Motion-preserved video restylization using point-tracks estimated from off-the-shelf trackers, while flexibly modifying visual appearance through multiple reference images;
    \item Mesh- or keypoint-driven video compositing and stylization using point-tracks derived from mesh vertices or keypoint detection;
    \item Camera control for both static and dynamic scenes using multiple reference images captured from different viewpoints and time steps, along with their associated static or dynamic point clouds.
    \end{enumerate*}
\end{itemize}

\section{Related work}

\label{sec:formatting}
\textbf{Video diffusion.}
Recent progress in video diffusion models~\cite{wan2025wan,kong2025hunyuanvideosystematicframeworklarge,genmo2024mochi,ray3_luma_ai,nvidia2025worldsimulationvideofoundation,hacohen2026ltx2efficientjointaudiovisual, hacohen2024ltxvideorealtimevideolatent} has dramatically improved the realism and temporal coherence of generated videos. Transformer-based diffusion frameworks~\cite{DiT} such as CogVideoX~\cite{CogVideoX} and Wan~\cite{wan2025wan} have pushed the boundaries of scalable video synthesis. However, while these models achieve high-quality video generation from text prompts, achieving precise controllability over what is generated and how it moves remains an open challenge. This motivates the development of reference and point-track conditioning beyond texts.

\noindent\textbf{Reference-to-video generation.}
Reference-to-video generation allows users to guide video synthesis with visual exemplars given as reference images, enabling more explicit control over the appearance and style of generated videos. Prior works~\cite{hu2025hunyuancustom, ye2025unic, cheng2025wan, xue2025standin} typically concatenate encoded reference images using pretrained VAEs along the temporal dimension and fine-tune self-attention layers of pre-trained video diffusion transformers for effective conditioning.
Our \method follows this paradigm but extends it by introducing point-track conditioning, which enables precise, spatially controlled reference insertion.

\noindent \textbf{Video-to-video generation.}
Another related line of work is video-to-video generation, which aims to apply semantic or stylistic edits~\cite{TokenFlow, zi2025senorita, li2025vfxmaster, mai2025easyv2, jiang2025vace, yatim2023spacetime, ju2025editverse}, or modify camera motion~\cite{bai2025recammaster, hu2025ex4d, yu2025trajcraft, luo2025camclonemaster, ren2025gen3c}, while maintaining temporal correspondences with the source video.
In contrast, we disentangle motion from appearance and train the model to condition on temporally coherent point-tracks together with temporally agnostic reference images. 
This design goes beyond standard video-to-video editing by enabling reference insertion, while still supporting conventional video editing through point-tracks extracted from the source video.

\noindent\textbf{Point-track-to-video generation.}
Point-track-conditioned video generation aims to control the motion of generated videos by conditioning on explicit point-track signals given as track prompts. 
Recent works~\cite{gu2025diffusion, zhang2025flextraj, geng2025motionprompting, zhang2024tora, zhang2025tora2, wang2025levitor, wang2025, chu2025wan, shin2025motionstream} have demonstrated the utility of track prompts for camera and object motion. However, most assume objects appear in the first frame or are completely specified by text prompts, which limits their ability to handle new content from different frames.
A key challenge lies in representing point-track identities for conditioning. Prior works employ random RGB values~\cite{burgert2025motionv2veditingmotionvideo} or sinusoidal embeddings~\cite{wang2025, burgert2025, shin2025motionstream}, which struggle with spatiotemporally discontiguous point correspondences—MotionV2V~\cite{burgert2025motionv2veditingmotionvideo} reports handling only 25 point-tracks optimally. First-frame coordinate encoding~\cite{zhang2025flextraj, gu2025das} maintains spatial smoothness but limits reference conditioning to the first frame. 
We instead propose learnable \textit{spatially-aware point-track embeddings} that encode spatial characteristics into unique identifiers, enabling the model to effectively associate and distinguish point-tracks.

\section{Method}

\subsection{Task formulation and inputs}
\textbf{Task.} Our goal is to create a multi-reference and point-track conditioned video generator. 
Specifically, given a set of reference images as input and user-specified 2D point-tracks, we aim to synthesize a video in which the visual elements from the reference images appear and evolve consistently over time according to the spatiotemporal conditions specified by the input point-tracks. 

\noindent \textbf{2D point-tracks anchored in reference images.} 
We establish correspondence between the generated video and the reference images using 2D point-tracks. 
Conventionally, a point-track defines the 2D flow of a point within a video sequence~\cite{cotracker}. We extend this definition by anchoring point-tracks across both generated video frames and input reference images, thus explicitly establishing point correspondences between them (see~\cref{fig:pipeline}). 
Formally, each point-track stores the 2D pixel coordinates $(x,y) \in [1, W] \times [1, H]$ in each video frame and the reference images. 
When a point-track is invisible (e.g., due to occlusion or leaving the frame) or filtered out (e.g., due to low confidence or unavailability in sky regions), its spatial coordinates are left undefined.
Our formulation supports both sparse and dense point-tracks, enabling flexible control. Sparse point-tracks allow the model to infer plausible motions based on neighboring track cues, providing soft guidance for motion generation. Dense point-tracks, in contrast, enforce strong spatial alignment and allow precise placement of reference objects within specific regions and frames. When point-track conditions appear only in the generated frames, they act as simple motion constraints. 
Conversely, point-tracks that appear only in the reference frames are discarded.  

\subsection{Training datasets and preprocessing}
\label{ssec:dataset_pre}

Due to the scarcity of video datasets with ground-truth point-track annotations, prior works~\cite{geng2025motionprompting,gu2025diffusion,zhang2025flextraj} annotate real video datasets using off-the-shelf point trackers~\cite{SpatialTracker,cotracker3}. 
However, these point trackers are predominantly trained on synthetic data such as Kubric~\cite{greff2022kubric}. 
This mismatch produces unreliable correspondences under strong camera motion or occlusions, resulting in noisy or incomplete point-tracks.

\noindent \textbf{Mixing static scene, synthetic, and real video datasets.} To address this limitation, we train on a combination of static scenes, synthetic dynamic scenes, and generic videos. 
We use the real-world static-scene dataset DL3DV~\cite{ling2024dl3dv} and synthetic static-scene dataset TartanAir~\cite{tartanair2020iros}, which provide ground-truth depth and camera pose, allowing us to compute accurate point-tracks.
We further use the dynamic synthetic dataset PointOdyssey~\cite{zheng2023point}, containing scenes with dynamic 3D meshes from which we derive per-vertex point-tracks to help disentangle camera motion from object motion. 
Finally, we use publicly available general-video datasets (e.g., OpenVidHD~\cite{nan2024openvid}, MiraData~\cite{ju2024miradatalargescalevideodataset},  OpenHumanVid~\cite{li2024openhumanvid}) annotated using the DELTA~\cite{ngo2024delta} tracker, which improves generalization to real-world footage. 
We captioned all videos using Gemini 2.5 Flash~\cite{gemini25flashimage2025} to enable joint training with text-conditioning.
Please see App.~\cref*{app:data_preprocess} for more details.

\noindent \textbf{Iterative point-track densification.} Previous methods sample dense tracks only from the first frame~\cite{geng2025motionprompting, shin2025motionstream, gu2025das, chu2025wan} or uniformly at random~\cite{zhang2025flextraj}, leading to poorly-tracked regions when objects rapidly undergo occlusion. 
We instead develop an iterative densification algorithm that detects and fills sparsely tracked regions, ensuring point-track coverage across the full scene. See Appendix~\cref*{app:iterative}.

\noindent \textbf{Training data construction.} 
Given a raw video and its point-tracks, we curate training data as triplets of: a) reference images, b) point-tracks, and c) corresponding target video frames to be generated. 
From each raw video, we extract a random sequence of $49$ consecutive frames as the target video. We then select $1-4$ reference frames \textit{from both within and outside this target sequence}.

\noindent \textbf{Reference image augmentation.} To ensure the model relies solely on point-tracks for spatial reasoning, we apply a random transform (zoom in/out, crop, resize, and translate) to each reference image. This guarantees that reference images are never spatially aligned with target video frames, preventing the model from learning trivial copy-paste shortcuts.

\noindent \textbf{Point-track augmentation.} Real-world point tracking is often incomplete due to occlusion and tracking failures. To improve robustness, we randomly drop track coordinates using three strategies: (1) \textit{visibility-based dropout}, which removes point-tracks that are not visible in the reference images; (2) \textit{spatial dropout}, which discards point-tracks that pass through randomly masked spatial regions; and (3) \textit{foreground overlay}, which overlays moving foreground masks to prevent the model from interpreting missing tracks as occlusions. We additionally vary the total number of point-tracks during training to improve robustness to diverse track densities at inference time. Detailed descriptions of these augmentations are provided in Appendix~\cref*{app:data_aug}.

\subsection{\method modeling and training}
\label{ssec:modeling}
\noindent \textbf{Encoding multiple reference images.} 
Following prior works~\cite{cao2025uni3c,bai2025recammaster,wang2024animatelcm,kant2025pippo}, reference images are encoded via the VAE and concatenated with noisy video tokens along the token dimension.
We introduce two modifications to distinguish reference tokens from those being denoised: (1) reference images are assigned RoPE~\cite{RoPE} positional indices starting from $100$, following~\cite{cao2025uni3c}; and (2) their timestep embedding is replaced with a learnable embedding initialized from timestep $0$, following~\cite{he2025unirelight}. This enables integration of reference images while preserving pretrained weights.

\noindent \textbf{Issue with randomly embedded point-tracks.} The key challenge in encoding point-tracks is enabling the model to associate a large number of point-tracks (up to $15{,}000$) between the target video frames to be generated and the reference frames. Prior methods~\cite{geng2025motionprompting, shin2025motionstream, burgert2025motionv2veditingmotionvideo} distinguish point-tracks by using randomly initialized embeddings. However, we find these spatially uncorrelated representations hinder accurate insertion of reference content into generated frames, and lead to slow convergence and poor generalization. 

\noindent \textbf{Spatially-aware point-track embedder.}
To address this, we encode point-tracks in a spatially-aware manner. As illustrated in the left part of \cref{fig:trajectory_architecture}, for each point-track, we augment its coordinates with frame indices $(i, x_i, y_i)$, pass them through a sinusoidal encoding and a shared coordinate MLP. Then, max-pooling over frames yields a single embedding that captures the point-track's spatial characteristics.
Here, we encode only generated-frame coordinates for the following reasons: (1) unlike reference frames, they exhibit motion continuity; (2) distinct tracks yield unique coordinate sequences, serving as natural identifiers; and (3) spatially proximate tracks have similar coordinate sequences, producing similar embeddings. This design provides spatially correlated, unique representations with minimal overhead.

\noindent \textbf{Point-track adapter.}
To inject pixel-space point-track embeddings into the video diffusion model, we aggregate them into the compressed patchified latent space ($4\times$ temporal, $16\times16$ spatial downsampling).
As illustrated in the middle part of \cref{fig:trajectory_architecture}, we divide the video into $4 \times 16 \times 16$ spatiotemporal blocks, each corresponding to a single latent token.
For each block, we collect all point-track embeddings within it.
Inspired by grid pooling in Point Transformer V2/V3~\cite{wu2022point, wu2024ptv3}, we aggregate via max pooling.
However, naive max pooling discards precise locations within each block. To address this, we concatenate each embedding with its relative position $(f', x', y') \in [0, 4) \times [0, 16) \times [0, 16)$ within the block and apply MLPs before pooling.
This preserves fine-grained positional information while operating over point-tracks in patchified latent space — not full pixel space.
Following prior work~\cite{he2025cameractrl2, ren2025gen3c}, we concatenate the adapter output channel-wise with video tokens after patchification and apply an MLP to recover the original channel dimension. We provide more details on each component in Appendix~\cref*{app:model}.

\noindent \textbf{Training details.}
We use \texttt{Wan 2.1-T2V-1.3B/14B} and \texttt{Wan-2.2 T2V-14B}~\cite{wan2025wan} as backbones, implemented in DiffSynth-Studio~\cite{diffsynthstudio2024}.
We train at $480 \times 832$ resolution with $49$ frames using AdamW~\cite{AdamW} (lr $1\text{e-}5$, weight decay $1\text{e-}3$) with 1000-step linear warmup for up to 24K iterations.
Text prompts and reference images are independently dropped with probability $0.15$ to enable joint classifier-free guidance at inference.
We train newly initialized weights from scratch and finetune self-attention and feed-forward layers in each DiT block, keeping cross-attention frozen.

\begin{table*}[!t]
    \caption{\textbf{Comparisons to baselines on DAVIS~\cite{davis2017}.} We compare against SOTA baselines. \textbf{Bold} indicates best results. The horizontal lines distinguish the different task settings. See App.~\cref*{tab:tapvid3d} for corresponding results on the TAPVid3D-ADT dataset.}
    \label{tab:main}
    \centering
    
    \scriptsize %
    \setlength{\tabcolsep}{4.5pt} %
    \renewcommand{\arraystretch}{1.1} %
    
    \begin{tabular}{lccccccc}
        \toprule
        \multirow{2}{*}{Method} 
        & \multirow{2}{*}{Backbone}
        & \multicolumn{2}{c}{Visual fidelity} 
        & \multicolumn{3}{c}{Reconstruction accuracy} 
        & \multicolumn{1}{c}{Motion fidelity} \\
        
        \cmidrule(lr){3-4}\cmidrule(lr){5-7}\cmidrule(lr){8-8}
        & 
        & FID $\downarrow$ 
        & FVD $\downarrow$
        & LPIPS $\downarrow$ 
        & PSNR $\uparrow$ 
        & SSIM $\uparrow$
        & EPE $\downarrow$ \\
        \midrule

        \multicolumn{8}{l}{\textbf{Dense tracks}} \\ %
        ATI~\cite{wang2025} & \texttt{Wan2.1 14B} & 41.69 & 504.9 & 0.369 & 13.93 & 0.448 & 16.20  \\
        DAS~\cite{gu2025diffusion} & \texttt{CogVideoX 5B} & 44.91 & 603.5 & 0.410 & 14.18 & 0.467 & 22.62 \\
        Tora~\cite{zhang2024tora} & \texttt{CogVideoX 5B} & 88.50 & 1382. & 0.537 & 12.28 & 0.351 & 42.37 \\
        GWTF~\cite{burgert2025} & \texttt{CogVideoX 5B} & 44.53 & 530.3 & 0.422 & 14.22 & 0.460 & 14.71 \\
        Wan-Move~\cite{chu2025wan} & \texttt{Wan2.1 14B} & 40.47 & 485.7 & 0.361 & 13.93 & 0.466 & 12.27 \\
        \rowcolor{gray!10}
        \textbf{\method (ours)} & \texttt{Wan2.1 1.3B} & 30.27 & 383.6 & 0.294 & 16.13 & 0.547 & 8.576\\
        \rowcolor{gray!10}
        \textbf{\method (ours)} & \texttt{Wan2.1 14B} & \underline{29.56} & \underline{341.5} & \underline{0.275} & \underline{16.49} & \underline{0.575} & \underline{7.958}\\
        \rowcolor{gray!10}
        \textbf{\method (ours)} & \texttt{Wan2.2 14B} & \textbf{28.00} & \textbf{322.8} & \textbf{0.265} & \textbf{16.86} & \textbf{0.589} & \textbf{7.709}\\
        \midrule 

        \multicolumn{8}{l}{\textbf{Mid-dense tracks (always visible on the first frame)}} \\
        ATI~\cite{wang2025} & \texttt{Wan2.1 14B} & 97.22 & 844.8 & 0.513 & 13.65 & 0.394 & 8.779 \\
        DAS~\cite{gu2025diffusion} & \texttt{CogVideoX 5B} & 45.98 & 624.4 & 0.388 & 14.23 & 0.481 & 16.19 \\
        Tora~\cite{zhang2024tora} & \texttt{CogVideoX 5B} & 114.5 & 1836. & 0.530 & 12.24 & 0.358 & 38.38 \\
        Wan-Move~\cite{chu2025wan} & \texttt{Wan2.1 14B} & 34.89 & 413.3& 0.338 & 14.86 & 0.507 & 6.834\\
        \rowcolor{gray!10}
        \textbf{\method (ours)} & \texttt{Wan2.1 1.3B} & 33.77 & 460.5 & 0.306 & 16.14 &  0.569 & \underline{3.829}\\
        \rowcolor{gray!10}
        \textbf{\method (ours)} & \texttt{Wan2.1 14B}  &  \textbf{29.42} & \textbf{340.5} & \underline{0.285} & \underline{16.38} & \underline{0.584} & \textbf{3.328}  \\
        \rowcolor{gray!10}
        \textbf{\method (ours)} & \texttt{Wan2.2 14B}       &  \underline{29.64} & \underline{363.0} & \textbf{0.281} & \textbf{16.53} & \textbf{0.598} & 3.932  \\
        \midrule 

        \multicolumn{8}{l}{\textbf{Sparse tracks (always visible on the first frame)}} \\
        ATI~\cite{wang2025} & \texttt{Wan2.1 14B} & 44.58 & 578.8 & 0.381 & 13.76 & 0.444 & 8.957\\
        DAS~\cite{gu2025diffusion} & \texttt{CogVideoX 5B} & 50.92 & 746.8 & 0.480 & 13.11 & 0.410 & 38.39 \\
        Tora~\cite{zhang2024tora} & \texttt{CogVideoX 5B} & 102.0 &  1702.&  0.546 & 12.42 & 0.359 & 39.22 \\
        Wan-Move~\cite{chu2025wan} & \texttt{Wan2.1 14B} & 38.49 & 496.6 & 0.371 &  14.14 & 0.459 & 8.316\\
        \rowcolor{gray!10}
        \textbf{\method (ours)} & \texttt{Wan2.1 1.3B} & 39.56 & 616.8 & 0.311 & 15.13 & 0.507 & \underline{5.155} \\
        \rowcolor{gray!10}
        \textbf{\method (ours)} & \texttt{Wan2.1 14B}  & \underline{34.06} & \textbf{485.3} & \underline{0.307} & \underline{15.39} & \underline{0.538} & \textbf{4.101} \\
        \rowcolor{gray!10}
        \textbf{\method (ours)} & \texttt{Wan2.2 14B} & \textbf{34.03} & \underline{500.0} & \textbf{0.302} & \textbf{15.59} & \textbf{0.547} & 5.173 \\
        \bottomrule
    \end{tabular}
\end{table*}

\section{Experiments}
\label{sec:experiments}
\noindent \textbf{Baselines and evaluation settings.} We compare against open-source state-of-the-art methods DiffusionAsShader~\cite{gu2025diffusion} (DAS), Go-with-the-Flow~\cite{burgert2025} (GWTF), and Tora~\cite{zhang2024tora} based on \texttt{CogVideoX-5B}~\cite{CogVideoX}, ATI~\cite{wang2025} and Wan-Move~\cite{wang2025} based on \texttt{WAN 2.1-14B}. 
Since we observe significant visual quality degradation in Tora and ATI with the large number of point-tracks, we subsample them up to $10$ and $40$ tracks, similar to the settings in \cite{shin2025motionstream}. 
For a comprehensive comparison, we show the results of \texttt{Wan-2.1-1.3B/14B} and \texttt{Wan-2.2-14B}, trained on the same architectures and datasets except the backbone. 
We always use the first frame as the reference image, matching the original problem settings in our baselines, and evaluate under three different point-track densities, all with first-frame reference:
\begin{itemize}    
    \item \textbf{Dense tracks:} We use 3K point-tracks that may start and end in any frame. For this task, we include GWTF as a baseline because of its capability to take dense optical flow as input. We also compare against DAS, ATI, and Wan-Move by removing point-tracks not starting from the first frame.
    \item \textbf{Mid-dense tracks (always visible on the first frame):} We use 512 point-tracks that all start from the first frame. This matches the conventional point-track-conditioned first-frame-to-video generation setting used in \cite{gu2025das, chu2025wan}. We exclude GWTF because it can see optical flows of content that appears in later frames.
    \item \textbf{Sparse tracks (always visible on the first frame):} We use only 32 point-tracks randomly sampled from the first frame. With this setting, ATI can now take all the 32 point-tracks without subsampling. We exclude GWTF because it can only be conditioned on dense optical flows.
\end{itemize}
We explain in Appendix~\cref*{app:exps_results} the exact settings under which we run the above baselines.

\noindent \textbf{Evaluation datasets and metrics.} Following prior works~\cite{geng2025motionprompting,shin2025motionstream}, we evaluate on the DAVIS 2017~\cite{davis2017} train-validation split of 77 videos with diverse camera motion, object motion, and occlusions. Furthermore, we use a random subset of 50 videos from TAPVid3D-ADT~\cite{tapvid3d}, which provides ground-truth point-tracks. Neither dataset is used during training. We group our evaluation metrics into three categories:
\begin{itemize}
    \item \emph{Visual fidelity:} Evaluated using FID and FVD, which compare the generated and reference videos and favor models close to the real-data distribution.
    \item \emph{Reconstruction accuracy:} Assessed via LPIPS, PSNR, and SSIM, which measure per-frame similarity between generated and ground-truth videos.
    \item \emph{Motion fidelity}: Measured as the L2 distance between visible input tracks and estimated tracks from generated videos using CoTracker3~\cite{cotracker3}. We call this metric endpoint error (EPE).
\end{itemize}

\begin{table*}[!t]
    \caption{\textbf{Ablation study of key design choices: spatially-aware point-track embeddings, point-track adapter, and hybrid training strategy.}
    Results are reported on DAVIS2017 and TAPVid3D-ADT, with the best performance highlighted in \textbf{bold} (per dataset).}
    \label{tab:ablation_traj}
    \centering
    \resizebox{\textwidth}{!}{%
    \begin{tabular}{lcccccccccccc}
        \toprule
        & \multicolumn{6}{c}{\textbf{DAVIS2017}} & \multicolumn{6}{c}{\textbf{TAPVid3D-ADT}} \\
        \cmidrule(lr){2-7}\cmidrule(lr){8-13}
        Ablation
          & FID $\downarrow$ & FVD $\downarrow$
          & LPIPS $\downarrow$ & PSNR $\uparrow$ & SSIM $\uparrow$
          & EPE $\downarrow$
          & FID $\downarrow$ & FVD $\downarrow$
          & LPIPS $\downarrow$ & PSNR $\uparrow$ & SSIM $\uparrow$
          & EPE $\downarrow$ \\
        \midrule

        \multicolumn{13}{l}{\textbf{Ablation on Point-track embeddings}} \\
        Random embeddings~\cite{shin2025motionstream, geng2025motionprompting}
          & 30.89 & 395.9 & 0.296 & 15.83 & 0.532 & 10.46
          & 47.52 & 375.5 & 0.319 & 16.99 & 0.649 & 6.141 \\
        \rowcolor{gray!10}
        Spatial-aware embeddings (\textbf{ours})
          & \textbf{28.93} & \textbf{347.6} & \textbf{0.288} & \textbf{16.19} & \textbf{0.548} & \textbf{7.983}
          & \textbf{45.74} & \textbf{331.5} & \textbf{0.302} & \textbf{18.10} & \textbf{0.662} & \textbf{4.691} \\
        \midrule

        \multicolumn{13}{l}{\textbf{Ablation on Point-track adapter}} \\
        Random sampling~\cite{chu2025wan, wang2025, burgert2025}
          & 32.34 & 431.5 & 0.306 & 15.86 & 0.517 & 10.97
          & 47.39 & 378.2 & 0.323 & 17.15 & 0.632 & 7.468 \\
        Max pool only
          & 31.93 & \underline{368.4} & 0.302 & 15.74 & 0.520 & \underline{9.539}
          & 47.51 & 382.9 & 0.312 & 17.41 & 0.642 & 6.356 \\
        Attention pooling \& rel.pos
          & \textbf{29.52} & \textbf{350.9} & \underline{0.295} & \underline{16.05} & \underline{0.535} & 9.552
          & \underline{45.09} & \underline{355.7} & \underline{0.300} & \underline{17.97} & \underline{0.663} & \underline{5.800} \\
        \rowcolor{gray!10}
        Max pool \& rel. pos (\textbf{ours})
          & \underline{30.29} & 377.4 & \textbf{0.286} & \textbf{16.11} & \textbf{0.555} & \textbf{8.851}
          & \textbf{44.41} & \textbf{335.2} & \textbf{0.296} & \textbf{17.98} & \textbf{0.668} & \textbf{4.677} \\
        \midrule

        \multicolumn{13}{l}{\textbf{Dataset Ablation (1.3B)}} \\
        Real video only
          & 33.18 & 460.0 & 0.315 & 15.52 & 0.522 & 10.59
          & 51.91 & 458.3 & 0.354 & 16.65 & 0.605 & 7.771 \\
        \rowcolor{gray!10}
        Real video, static scene \& synthetic (\textbf{ours})
          & \textbf{31.98} & \textbf{399.0} & \textbf{0.307} & \textbf{15.84} & \textbf{0.526} & \textbf{8.801}
          & \textbf{48.24} & \textbf{419.8} & \textbf{0.328} & \textbf{16.69} & \textbf{0.636} & \textbf{5.285} \\
        \bottomrule
    \end{tabular}%
    }
\end{table*}

\begin{figure*}[tb]
    \centering
        \captionsetup{type=figure}
        \includegraphics[width=\textwidth]{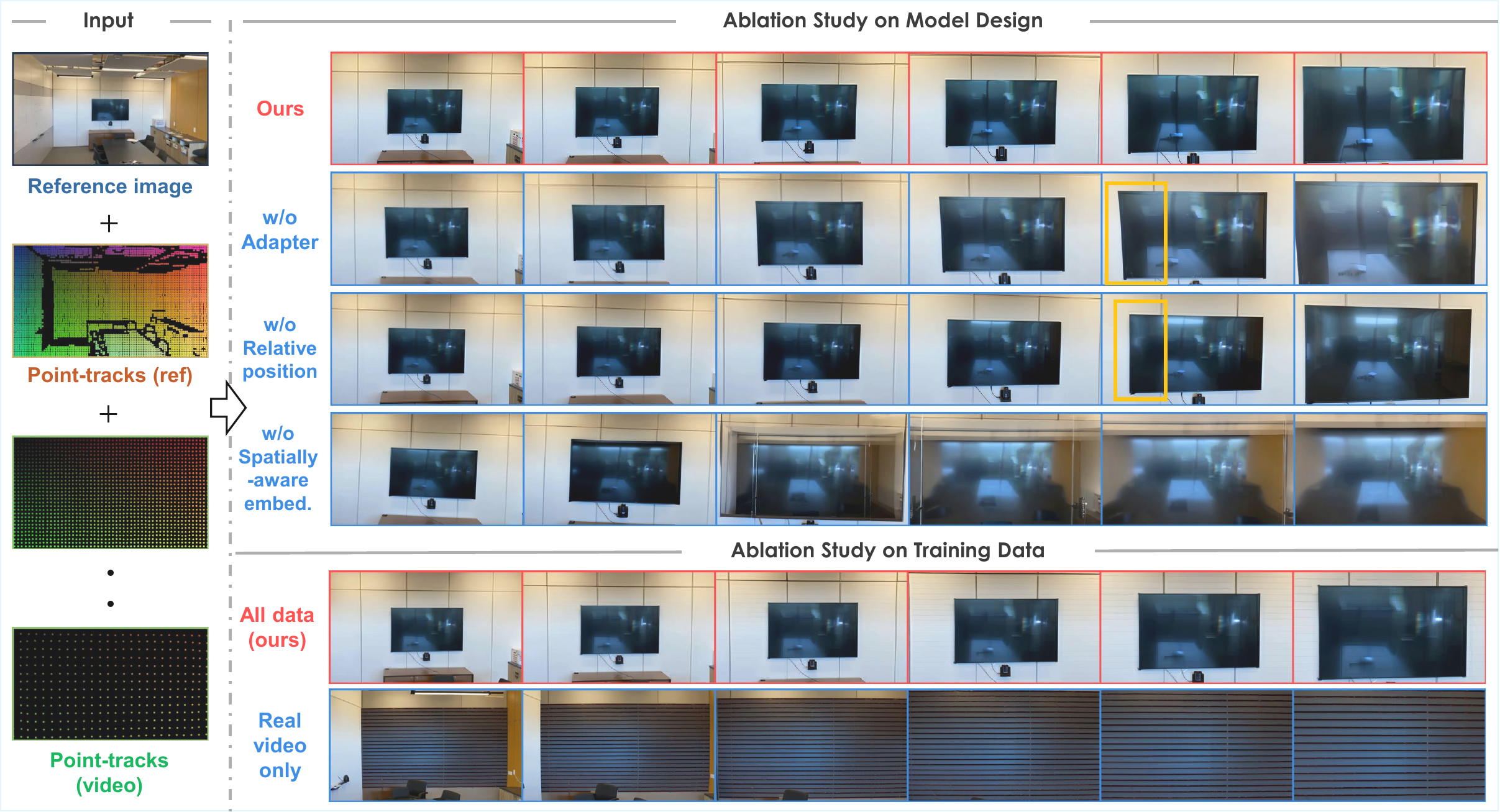}
        \captionof{figure}{\textbf{Qualitative ablation analysis.} Although the task is simply to generate zoomed-in videos of the TV given reference images of a zoomed-out conference room, removing our spatially-aware point-track embeddings, point-track adapter, relative position injection, or hybrid training data strategy results in geometric distortions and severe motion-following failures. In contrast, our full model accurately adheres to the prescribed point tracks and produces correct zoomed-in TV videos. Additional results are available on the supplementary webpage.}
    \label{fig:ablation_sup}
\end{figure*}

\subsection{Quantitative comparisons}

\textbf{Baseline comparisons.} In~\cref{tab:main}, we compare baselines against our method. 
As demonstrated in~\cref{tab:main}, our \method outperforms all the baselines, and effectively utilizes point-tracks starting and ending at arbitrary frames. Baselines such as ATI, Wan-Move, and DAS cannot leverage point-tracks that do not appear in the first frame, resulting in a significant loss of motion conditioning information. While GWTF uses dense optical-flow, optical-flow cannot reliably track long-range movements, especially under occlusion.

\method continues to outperform baselines for mid-dense and sparse tracks, even with all the point-tracks being visible on the first frame and thus allowing for a fair comparison to strong baselines such as Wan-Move. 
We attribute our superior performance to differences in how motion conditioning is handled. Specifically, prior methods downsample point-track inputs either through the VAE encoder (e.g., DAS) or via na\"ive point-track down/subsampling strategies (e.g., GWTF, ATI, and Wan-Move). These approaches tend to lose fine-grained motion details, leading to weaker motion conditioning.

Moreover, the performance gap between our method and the baselines becomes larger on challenging datasets such as TAPVid3D-ADT~(Appendix~\cref*{tab:tapvid3d}) with dense tracks.
We hypothesize that because previous methods are trained exclusively on real video datasets with noisy point-track labels, their ability to follow motion precisely is weakened. This demonstrates the effectiveness of our model architecture and real-synthetic data-mixing strategy.  

\begin{table}[t]
    \caption{\textbf{Ablation on varying numbers of reference images.} We report video quality, reconstruction quality, and motion accuracy for different reference-image settings.
    A keyframe means that the reference image appears exactly in the generated frames, whereas non-keyframe images do not match any target video frame.
    Please note that we use Wan2.1 14B (18,000 iterations) for this comparison. \textbf{Bold} indicates best results.}
    \centering
    \scriptsize
    \begin{tabular}{lcccccc}
        \toprule
        & \multicolumn{2}{c}{{Visual fidelity}} 
          & \multicolumn{3}{c}{{Reconstruction accuracy}} 
          & \multicolumn{1}{c}{{Motion fidelity}} \\
        \cmidrule(r){2-3}\cmidrule(lr){4-6}\cmidrule(l){7-7}
        Ablation
          & FID $\downarrow$ & FVD $\downarrow$
          & LPIPS $\downarrow$ & PSNR $\uparrow$ & SSIM $\uparrow$
          & EPE $\downarrow$ \\
        \midrule

        \multicolumn{7}{l}{\textbf{Varying Keyframe Selection Strategy}} \\ %
        \hspace{3mm} First
          & 27.83 & 346.4 & 0.278 & 16.18 & 0.554 & 7.776 \\
        \hspace{3mm} Middle
          &  23.78  & 315.5 & 0.242 & 17.22 & 0.586 & 7.651 \\
        \hspace{3mm} Last 
          & 39.98 & 466.1 & 0.356 & 14.51 & 0.497 & 9.047  \\
        \hspace{3mm} First-Last 
          & 19.60 & 233.3 & 0.209 & 18.09 & 0.615 & 7.198 \\
        \hspace{3mm} 4-Uniform 
          & \textbf{14.89} & \textbf{188.7} & \textbf{0.170} & \textbf{19.50} & \textbf{0.647} & \textbf{7.027}\\
        
        \midrule %
        
        \multicolumn{7}{l}{\textbf{Non-keyframe reference images}} \\
        \hspace{3mm} RandomCrop 
          & 36.19 & 429.9 & 0.337 & 14.91 & 0.502 & 8.074 \\
        \bottomrule
    \end{tabular}
    \label{tab:keyframe}
    \vspace{-5pt}
\end{table}

\noindent \textbf{Using multiple keyframe and reference images.} To evaluate our model's multi-reference conditioning capabilities, we test under several settings: (a) single frame (first, middle, last), (b) first and last frames, and (c) four frames (first, last, and two middle). 
 As shown in Tab.~\ref{tab:keyframe}, our model performs stably regardless of which references are provided, with consistent improvements as the number of references increases. We further validate flexibility by constructing an evaluation set where reference frames lie outside the generation interval, with random cropping and resizing applied. 
 Our method maintains robust performance, confirming its ability to leverage reference images even without exact spatial alignment in the target frames.

\subsection{Ablation study}
We study the effectiveness of our key designs: spatially-aware point-track embedder, point-track adapter, and the hybrid training dataset strategy, with quantitative results in~\cref{tab:ablation_traj}, and visual comparisons in~\cref{fig:ablation_sup} and the supplementary webpage.

\begin{figure*}[!tb]
    \centering
    \captionsetup{type=figure}
        \includegraphics[width=\textwidth]{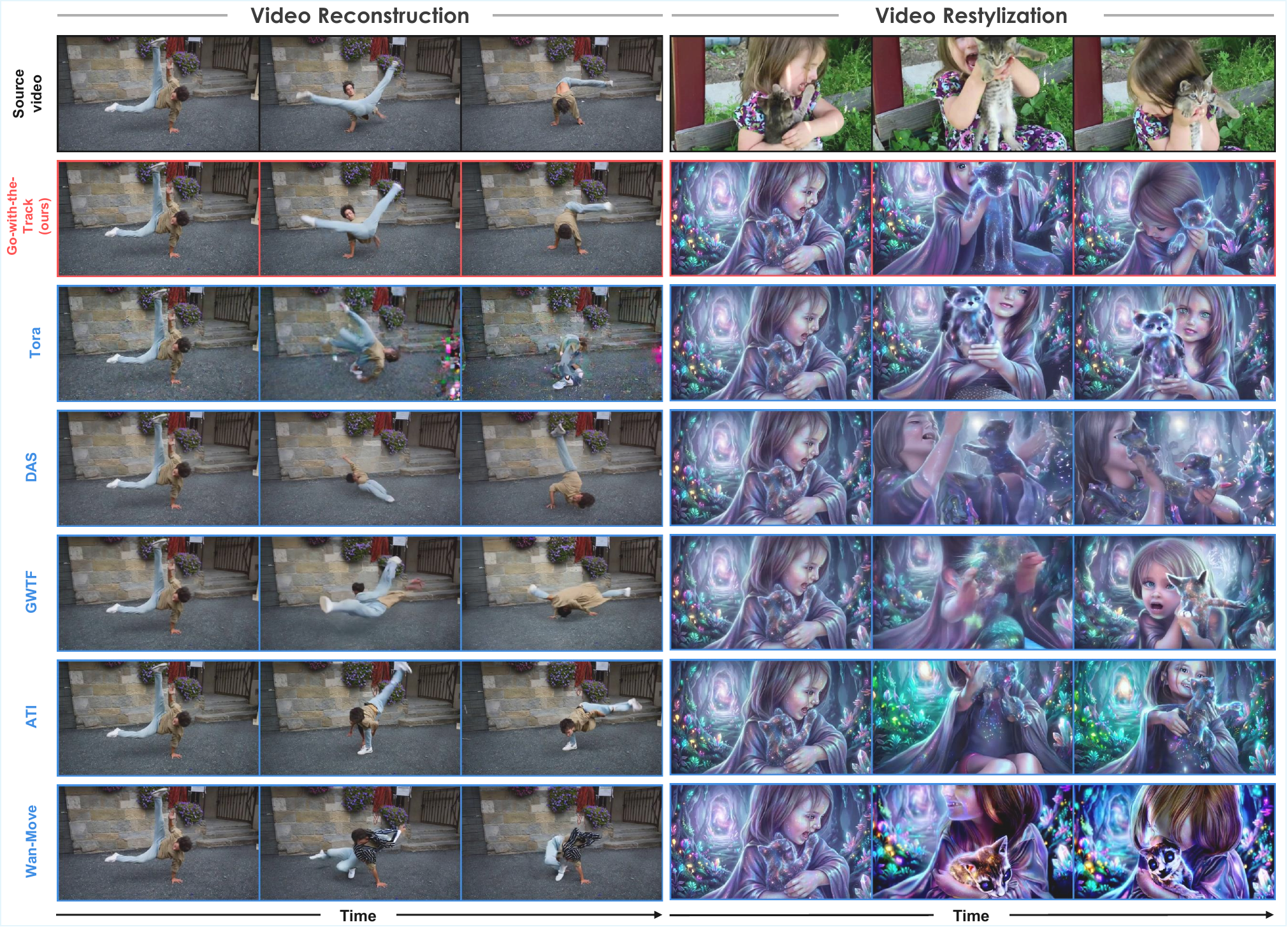}
        \captionof{figure}{\textbf{Qualitative comparisons to baselines on video reconstruction and video restylization.}  (Left) Given the first frame and point-tracks extracted from the source video, each method attempts to reconstruct the source video. (Right) Given the stylized first frame and point-tracks, each method generates
motion-preserved restylized videos. 
        As evident, Go-with-the-Track better preserves the source motion while adhering to the appearance of the source/stylized first frame.
        Additional results are provided in the supplementary webpage.}

    \label{fig:supp_davis}
\end{figure*}

\noindent \textbf{Random vs spatially-aware point-track embeddings.} A key design choice in our framework is the use of spatially-aware point-track embeddings rather than random embeddings. 
Previous motion control methods~\cite{MotionCtrl, shin2025motionstream, burgert2025motionv2veditingmotionvideo} assign random vectors to distinguish individual point-tracks, which provide no inherent spatial cues for the model to leverage. 
In contrast, our embeddings encode the coordinates of each point-track, providing the model with explicit spatial priors that help distinguish and locate point-tracks across video frames and reference images. 
As shown in~\cref{tab:ablation_traj}, replacing our spatially-aware embeddings with random vectors of the same dimensionality leads to clear performance drops in both motion controllability and visual fidelity. 
This indicates that when point-tracks carry spatially meaningful representations, the model can better reason about the correspondence between point-tracks and their associated regions in the reference image, thereby improving visual coherence and motion controllability. Appendix~\cref*{fig:supp_pca} presents a PCA visualization comparing random embeddings and our point-track embeddings. The results show that our embeddings exhibit clear spatial correlations.

\noindent \textbf{Point-track adapter.} We design our adapter that preserves motion details without excessive computational overhead, unlike na\"ive downsampling or VAE-based approaches~\cite{shin2025motionstream,chu2025wan}. We compare: (1) \emph{random sampling} of a single point-track per spatiotemporal block~\cite{chu2025wan, wang2025}, (2) \emph{max pooling only} without relative position injection, and (3) \emph{attention pooling} instead of max pooling. As shown in~\cref{tab:ablation_traj}, random sampling performs poorly; max pooling improves slightly, but concatenating relative positions before pooling achieves best results by preserving spatial location information. Max pooling outperforms attention pooling, consistent with design decisions in PointNet~\cite{PointNet} and recent point transformers~\cite{wu2022point, wu2024ptv3}. We hypothesize that max pooling is better suited to this setting because it preserves salient per-track embedding values corresponding to distinct identities, while avoiding the smoothing effects introduced by weighted averaging in attention-based aggregation.

\noindent \textbf{Hybrid training dataset strategy.}
To validate the importance of training on datasets with precise point tracking annotations, we remove the static-scene and synthetic datasets (Tartanair~\cite{tartanair2020iros}, DL3DV~\cite{ling2024dl3dv}, and PointOdyssey~\cite{zheng2023point}) and train solely on our real-video dataset (e.g., OpenVidHD~\cite{nan2024openvid}, MiraData~\cite{ju2024miradatalargescalevideodataset}, OpenHumanVid~\cite{li2024openhumanvid}) and show quantitative results in~\cref{tab:ablation_traj}. 
We see a clear degradation in the motion fidelity metric (EPE), as well as a drop in reconstruction accuracy, in both the general (DAVIS 2017) and challenging (TAPVid3D-ADT) evaluation datasets.

\vspace{-5pt}
\subsection{Qualitative comparisons}
\textbf{Baseline comparisons.} \cref{fig:supp_davis} and our supplementary webpage present qualitative comparisons between \method and the baselines on video reconstruction and restylization tasks, using the first frame as the reference frame. The motion generated by \method closely aligns with the source video, while better preserving spatial structure and subject identity of the first frame than the baselines.

\newcommand{\best}[1]{\textbf{#1}}
\newcommand{\second}[1]{\underline{#1}}

\begin{table}[t]
    \caption{\textbf{User Study.} We show preference (\%) per question for all methods across 45 participants who answered 90 questions each.}
    \centering
    \scriptsize
    \setlength{\tabcolsep}{2pt}
    \begin{tabular}{lccc}
        \toprule
        Method & Q1: Motion following & Q2: Subject preservation & Q3: Overall quality\\
        \midrule
        ATI~\cite{wang2025} & 15.7 & 17.2 & \second{18.6} \\
        GWTF~\cite{burgert2025} & \second{17.7} & \second{17.5} & 16.0 \\
        DaS~\cite{gu2025diffusion} & 10.5 & 10.8 & 11.2 \\
        Tora~\cite{zhang2024tora} & 10.0 & 11.1 & 9.9 \\
        \rowcolor{gray!10}
        \textbf{\method (ours)} & \best{46.2} & \best{43.5} & \best{44.3} \\
        \bottomrule
    \end{tabular}
    \label{tab:user_study}
\vspace{-15pt}
\end{table}

\noindent\textbf{User study.} We recruited 45 participants to evaluate 30 randomly selected videos based on motion following, subject preservation, and overall quality. As shown in Tab.~\ref{tab:user_study}, our method is preferred more than twice as often as the second-best method, demonstrating its effectiveness in motion control, identity preservation, and visual quality. User study details are provided in Appendix \cref*{qual_detail_supp}.

\vspace{-5pt}
\subsection{Applications}

\method supports flexible point-track and reference conditioning, enabling a broad range of applications. Representative examples are shown in Fig.~\ref{fig:teaser}. Additional visual results and implementation details can be found in Appendix~\cref*{sec:sup_app_detail} and on our supplementary webpage. 

\noindent\textbf{Video restylization.} Given a source video, we first extract point-tracks using an off-the-shelf tracker~\cite{ngo2024delta}. To construct subject-specific restylized references, we select frames from the video and apply SAM3~\cite{carion2025sam3segmentconcepts} to segment and crop the target objects, which are subsequently stylized with Nano-Banana~\cite{gemini25flashimage2025}. As illustrated in Fig.~\ref{fig:teaser} (first part), our method preserves complex motion patterns while faithfully adhering to the reference appearance.

\noindent\textbf{Mesh-driven compositing and stylization.} Given an animated mesh, we render a reference frame, stylize it using \cite{gemini25flashimage2025}, and extract vertex trajectories under a specified camera path. As shown in Fig.~\ref{fig:teaser} (second part), our method transfers the style while faithfully preserving the original animation, including human motion and scene interactions.

\noindent\textbf{Keypoint-driven compositing.} Given a human-centric video and a reference image, we apply off-the-shelf facial and human-body keypoint detectors~\cite{giebenhain2025pixel3dmm, yang2026sam3dbody} to derive a sequence of point correspondences between generated frames and reference images. As shown in Fig.~\ref{fig:teaser} (third part), by treating keypoints as a point-track, we can transfer the appearance of the reference human to the source video. 

\noindent\textbf{Camera control in static and dynamic scenes.} For static scenes, we use $\pi^3$~\cite{wang2025pi} to reconstruct 3D point clouds and recover camera poses, then project points onto target camera trajectories. For dynamic scenes, we perform 3D tracking using DELTA~\cite{ngo2024delta} with iterative point-track densification (Appendix \cref*{app:iterative}), estimate per-frame cameras via $\pi^3$, and reproject tracks under custom camera trajectories—excluding points marked invisible to avoid noisy reprojections. For camera retargeting, we uniformly sample four frames from the source video as reference images. As shown in Fig.~\ref{fig:teaser} (fourth and fifth parts), \method supports various camera paths including spiral motions and multi-view interpolation, correctly warping references even to significantly different viewpoints.

\vspace{-10pt}

\section{Conclusion}

\label{sec:conclusion}
We introduce \method, a unified framework that combines point-track-driven motion control with multi-reference compositing for precise appearance and motion control in video generation. By representing spatio-temporal relationships with dense point-tracks, our method provides reliable control over when and where reference elements appear and how they move. We propose a spatially-aware point tracking ID embedding with a lightweight encoder that effectively maps track identities into the patchified latent space. Extensive experiments demonstrate superior controllability and video quality, while qualitative examples highlight flexibility across mesh stylization, motion transfer, camera control, and track-level editing.

\noindent \textbf{Limitations.} Performance depends on point-track quality; fast motion or incorrect visibility estimates can introduce temporal artifacts. Track spatial resolution limits fine-grained precision for detailed contact or subtle motions, and output quality depends on the underlying video diffusion model.

\begin{acks}
We would like to thank Pablo Salamanca, Simon Su, and Nick Abrahan for their technical support; Nhat Phong Tran and Mingming He for their production support; Jeffrey Shapiro, Ritwik Kumar, and Hossein Taghavi for their executive support; Jennifer Lao and Lianette Alnaber for their operational support. Koichi acknowledges support from the Clarendon Scholarship.
\end{acks}

\bibliographystyle{ACM-Reference-Format}
\bibliography{main}

\appendix
\clearpage
\appendix

\section{Dataset preprocessing}
\label{app:data_preprocess}

\textbf{Dataset filtering on real video datasets.}
To train our model on real video datasets, we collect publicly available general-video datasets (e.g., OpenVidHD~\cite{nan2024openvid}, MiraData~\cite{ju2024miradatalargescalevideodataset}, OpenHumanVid~\cite{li2024openhumanvid}), exposing the model to various types of videos. 
However, we observe substantial variations in content quality across datasets. Some videos contain frequent scene cuts, overlaid subtitles, or minimal motion, which may negatively impact training. To ensure high-quality training data, we apply a series of automated filtering procedures. Specifically, we use OCR-based filtering~\cite{cui2025paddleocr30technicalreport, cui2025paddleocrvlboostingmultilingualdocument} to remove videos with prominent text overlays, scene cut detection~\cite{soucek2020transnetv2} to exclude videos with abrupt transitions, and optical-flow-based motion filtering~\cite{farneback} to discard videos with insufficient motion.
After this preprocessing pipeline, we obtain approximately 500K curated high-quality videos for training.

\noindent \textbf{Point tracking.} To obtain point-tracks for real videos without ground-truth annotations, we employ the DELTA point tracker~\cite{ngo2024delta}. Because DELTA requires metric depth as input, we first estimate disparity depth using VideoDepthAnything~\cite{chen2025videodepthanythingconsistent}, which better preserves fine-grained foreground geometry. 
Next, we estimate metric depth with UniDepthV2~\cite{unidepth} and align the disparity map to metric depth via least-squares optimization, resolving the scale ambiguity. All point tracking is performed at a resolution of $384 \times 512$, following DELTA’s default configuration.

\noindent \textbf{Dataset captioning.}
In addition to reference images and point-track controls, textual captions provide complementary contextual information that cannot be fully captured by tracks or visual references alone, thereby offering stronger guidance for video generation.
We manually evaluated a subset of captions generated by Qwen2.5~\cite{yang2024qwen25} and Gemini 2.5 Flash~\cite{gemini25flashimage2025}. Based on this comparison, we found Gemini 2.5 Flash to produce more accurate and faithful descriptions, and therefore adopt it for all experiments.
To caption each video, we use the following prompt:

\textit{Write a single-paragraph caption (under 200 words) that describes the video in rich visual detail. If humans are present, describe their appearance (e.g., clothing, hairstyle, physical features), motion (e.g., gestures, walking, running), and any interactions with other people, objects, or the environment. If no humans appear, focus on the appearance and motion of animals, vehicles, or other active elements. Always describe the scene or background in detail, including location, objects, environment, and overall atmosphere. Include lighting information, such as brightness, color temperature, shadows, and whether the light is natural or artificial. Describe the camera work, including framing (e.g., close-up, wide shot), angle (e.g., low, high, overhead), and movement (e.g., static, panning, tracking). Use full sentences in a single paragraph without bullet points. Be as specific and informative as possible while avoiding redundancy.}

\noindent \textbf{Dataset Mixing Ratios.} In practice, we combine real videos, DL3DV~(real, static), PointOdyssey~(synthetic, dynamic), and TartanAir~(synthetic, static) with a sampling ratio of 11:3:3:3. This results in an approximate 3:7 synthetic-to-real data ratio, a 3:7 static-to-dynamic scene balance, and maintain a 9:11 ratio between ground-truth point-tracks and off-the-shelf point-tracker-estimated point-tracks.

\section{Iterative densification of point tracks}
\label{app:iterative}
\begin{algorithm}[t!]
    \scriptsize
    \caption{Iterative resampling of point queries}
    \label{alg_iter}
    
    \SetKw{KwTo}{to}
    \SetKw{KwStep}{step}
    \SetInd{0.25em}{0.55em}
    \SetKwInOut{Input}{input}
    \SetKwInOut{Output}{output}
    \SetKw{Return}{return}

    \Input{%
      video $V$ with $F$ frames of size $H \times W$;\\
      maximum number of query points $N$; iterations $T$; patch size $C$;
    }
    \Output{final query set $Q$}

    \BlankLine
    \tcp{Initialize with uniformly random sampled query points}
    $n_{\mathrm{per}} \gets \lfloor N/T \rfloor$ \;
    $Q \gets \mathrm{UniformSample}(V, n_{\mathrm{per}})$ \;

    \For{$t \gets 1$ \KwTo $T$}{
        \tcp{Apply point tracker}
        $\mathrm{Track} \gets \mathrm{DELTA}(V, Q)$ \;
        
        \tcp{(2) Enumerate empty patches on a $C \times C$ grid}
        $Q_{\mathrm{new}} \gets [ \ ]$ \;
        
        \For{$f \gets 1$ \KwTo $F$}{
            \For{$y \gets 1$ \KwTo $H$ \KwStep $C$}{
                \For{$x \gets 1$ \KwTo $W$ \KwStep $C$}{
                    \If{$\mathrm{Track}[f, y:y+s_h, x:x+s_w]$ is empty}{
                        $c_x \gets x + \lfloor C/2 \rfloor$; $c_y \gets y + \lfloor C/2 \rfloor$ \;
                        $Q_{\mathrm{new}}.\mathrm{append}((f, c_x, c_y))$ \;
                    }
                }
            }
        }

        \tcp{(3) Add up to $N/T$ new queries}
        $Q \gets Q \cup \mathrm{RandomSubset}(Q_{\mathrm{new}}, n_{\mathrm{per}})$ \;
    }
\end{algorithm}

\noindent \textbf{Issue with uniformly sampled point-track queries.} A point-tracker (\eg DELTA) takes as input a video and a list of tuples $(u,v,t)$ that specify the pixel locations of points to be tracked in the 3D space-time volume, called query points.

Prior works~\cite{gu2025diffusion,geng2025motionprompting} either sample query points exclusively from the first frame or uniformly across all frames~\cite{zhang2025flextraj}.  
We observe that both these approaches suffer from unwanted sparse regions, particularly for small objects or background elements that appear only briefly.
For example, sampling 2,500 query points uniformly over an 81-frame video (Wan's context length) results in approximately 30 query points per frame.   
For a 480P video, thus randomly sampling 30 point-queries on the 2D image grid leads to poor spatial coverage, failing to capture fine-grained motion details while over-sampling query points to large or dominant objects.
    
\noindent\textbf{Iterative densification of point tracks.} To mitigate this issue, we propose an iterative track-query resampling scheme (Alg.~\ref{alg_iter}) that fills sparse regions while avoiding oversampling. Let $T$ be the number of iterations and $N$ the total number of point queries. At each iteration, we partition each frame into a $C \times C$ grid (with $C = 20$), identify patches without track points, randomly select up to $N/T$ empty patches, and sample one new query point per selected patch. We then rerun the tracker to update all point-tracks. Repeating this procedure progressively reduces sparsity and improves spatial coverage, capturing fine-grained motion in small or short-lived regions without oversampling dense areas. 

\noindent\textbf{Iterative point-track densification.} Unlike naive frame-by-frame resampling, which would require running the tracker $F$ times (once per frame), our iterative strategy achieves dense coverage efficiently with a small number of iterations. Empirically, setting the number of iterations $T = 5$ and the total number of track queries $N = 3000$ effectively eliminates sparse regions while remaining computationally feasible. The algorithm can run on a Tesla T4 GPU (15 GB memory) within practical preprocessing time, enabling efficient large-scale dataset preparation. 

\begin{figure*}[!t]
    \centering
        \captionsetup{type=figure}
        \includegraphics[width=\textwidth]{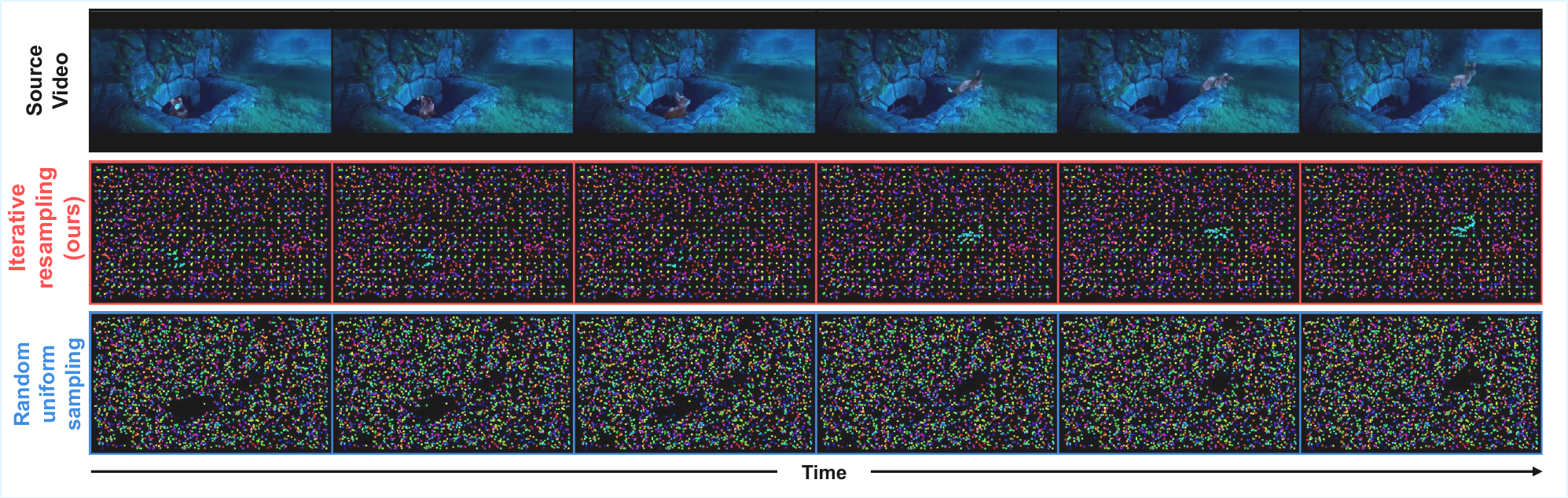}
        \captionof{figure}{\textbf{Iterative point-track resampling vs. random uniform sampling.} 
Visual comparison of detected point-tracks obtained using our iterative resampling strategy (\cref{alg_iter}) and uniform random sampling of point queries over the video frames. Our iterative resampling produces denser and more uniformly distributed point-tracks, achieving better spatial coverage with reduced sparsity. Please refer to the supplementary webpage for the corresponding video.}
    \label{fig:algo_iter}
\end{figure*}

\noindent\textbf{Visual comparison.} In \cref{fig:algo_iter}, we present a visual comparison of the detected point tracks obtained using our iterative resampling technique versus uniform sampling of point queries over space and time in the video. 
As evident from the figure, the iteratively resampled point tracks provide substantially better coverage and exhibit reduced sparsity, particularly for small objects undergoing rapid motion across video frames.

\noindent\textbf{Static-scene datasets.} Static scene datasets such as DL3DV~\cite{ling2024dl3dv} and TartanAir ~\cite{tartanair2020iros} naturally exhibit sparse regions where few or no tracks are available due to missing depth annotations. Rather than correcting this sparsity, we treat it as an implicit form of data augmentation that further enhances the model’s robustness to incomplete supervision.

\begin{figure*}[htbp] %
  \centering
  \includegraphics[width=\textwidth]{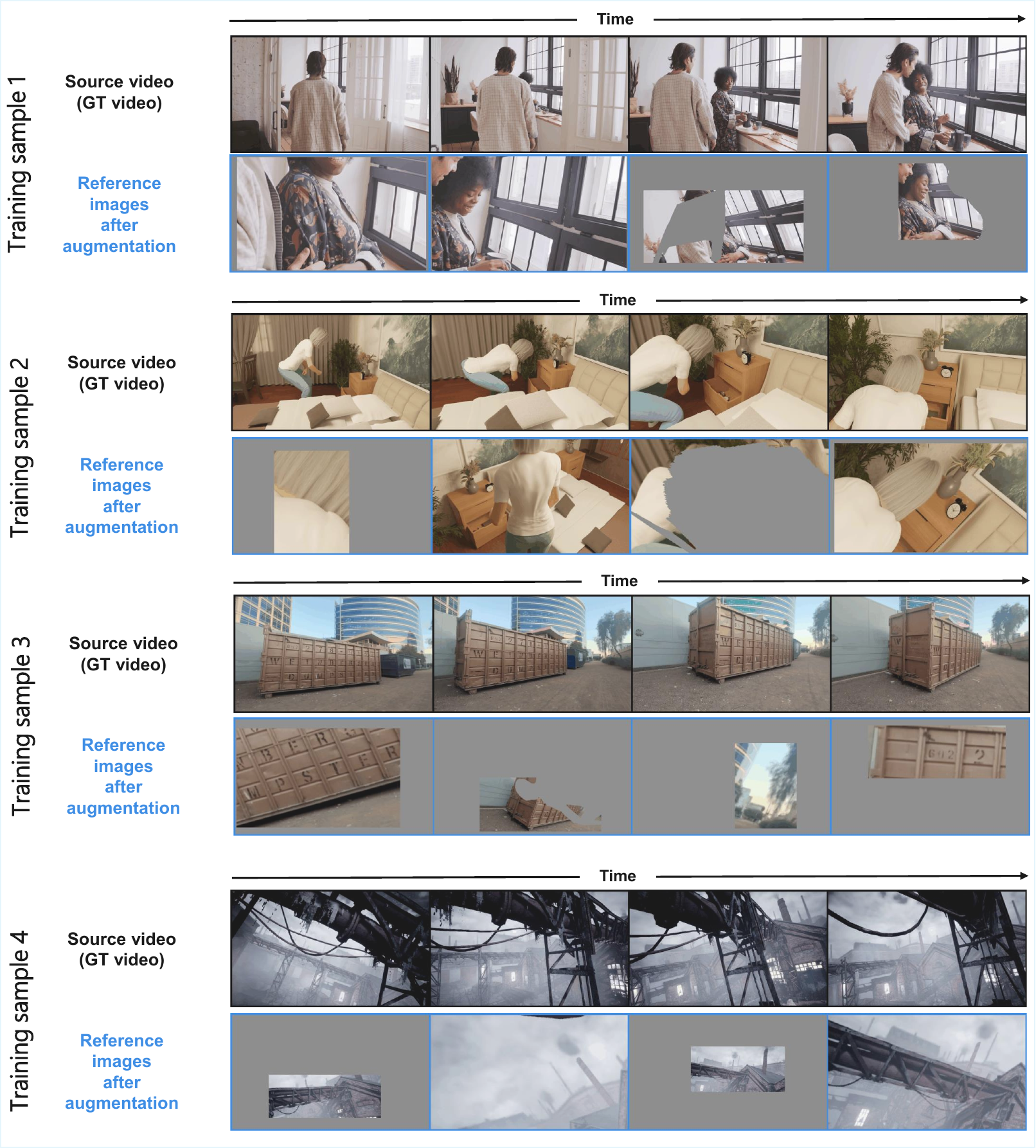}
  \caption{\textbf{Examples of augmented training samples.} Visualization of training samples after applying the data augmentation strategies described in \cref{app:data_aug}. The examples highlight the diversity of motion patterns and reference image variations. For clarity, point-track conditions are omitted. Please refer to the supplementary webpage for additional examples.}
  \label{fig:train_data_}
\end{figure*}

\section{Details of point-track augmentation.}
\label{app:data_aug}

To improve robustness to the various types of sparse or incomplete point tracks, which is common in real-world scenarios due to occlusion or tracking failures, we randomly mark a subset of track coordinates as undefined using three complementary strategies:

\begin{itemize}

    \item \textit{Visibility-based dropout:} We discard point-tracks that are not visible in the reference images. This simulates scenarios in which motion is specified only for elements present in the reference images (e.g., conventional point-track-conditioned first-frame-to-video generation). Consequently, for content that appears only in the generated frames, the model must infer motion and appearance without explicit point-track guidance.

    \item \textit{Spatial dropout:} We apply a random 2D mask to a randomly selected frame and remove all point-tracks passing through the masked region across the entire sequence. This mimics missing point-tracks in geometrically ambiguous areas (e.g., sky) and enables flexible mixing of sparse and dense conditioning during inference.

    \item \textit{Foreground overlay:} We observe that the model tends to interpret untracked regions as being occluded by unseen foreground objects, often leading to hallucinated content~\cite{geng2025motionprompting}. To mitigate this effect, we overlay moving foreground masks from YouTube-VOS~\cite{xu2018youtubevoslargescalevideoobject} onto training videos without providing associated tracks. This encourages the model to learn that missing point-tracks do not necessarily imply occlusion.

\end{itemize}

These strategies are applied with probabilities of $7\%$, $7\%$, and $15\%$ respectively, as we observe that more aggressive augmentation destabilizes training and degrades motion controllability. 
Finally, we randomly subsample the number of point tracks to between $256$ and $16{,}384$, exposing the model to a wide range of point-track densities. 
Examples of the augmented training samples are shown in \cref{fig:train_data_} and on our supplementary webpage.

\begin{figure*}[htbp] %
  \centering
  \includegraphics[width=\textwidth]{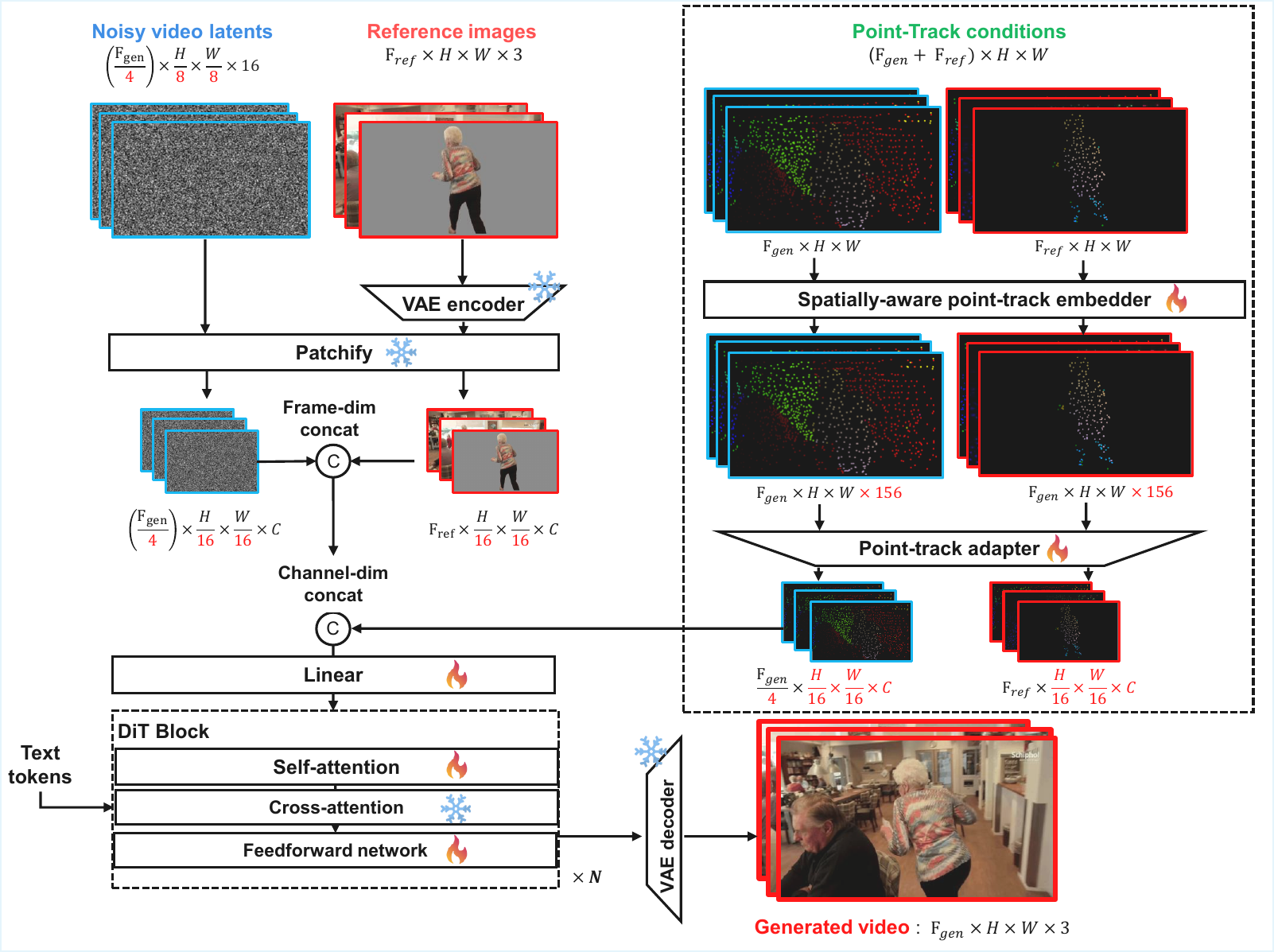}
  \caption{\textbf{Details of the \method pipeline.} \method is built upon the pre-trained video diffusion models \texttt{Wan2.1/2.2-T2V}~\cite{wan2025wan}. For clarity, timestep conditioning is omitted from the diagram. \textbf{Top left:} VAE-encoded reference images and noisy target frames are concatenated along the temporal dimension. \textbf{Right:} Both reference and generated frames are conditioned on point-tracks. The point-tracks are first encoded into spatially-aware point-track embeddings, which are then downsampled using our lightweight point-track adapter to align with the resolution of the patchified latent space. \textbf{Bottom left:} The resulting point-track embeddings are concatenated channel-wise with the patchified tokens of the noisy video and reference frames, followed by a linear projection layer, and then processed by partially fine-tuned DiT blocks.}
  \label{fig:overall_supp}
\end{figure*}

\section{Model architecture details}
\label{app:model}
In this section, we detail our model architectures as introduced in \cref{ssec:modeling}. 
A detailed overview of the complete pipeline is illustrated in \cref{fig:overall_supp}.

\begin{figure}[!t] %
  \centering
  \includegraphics[width=\columnwidth]{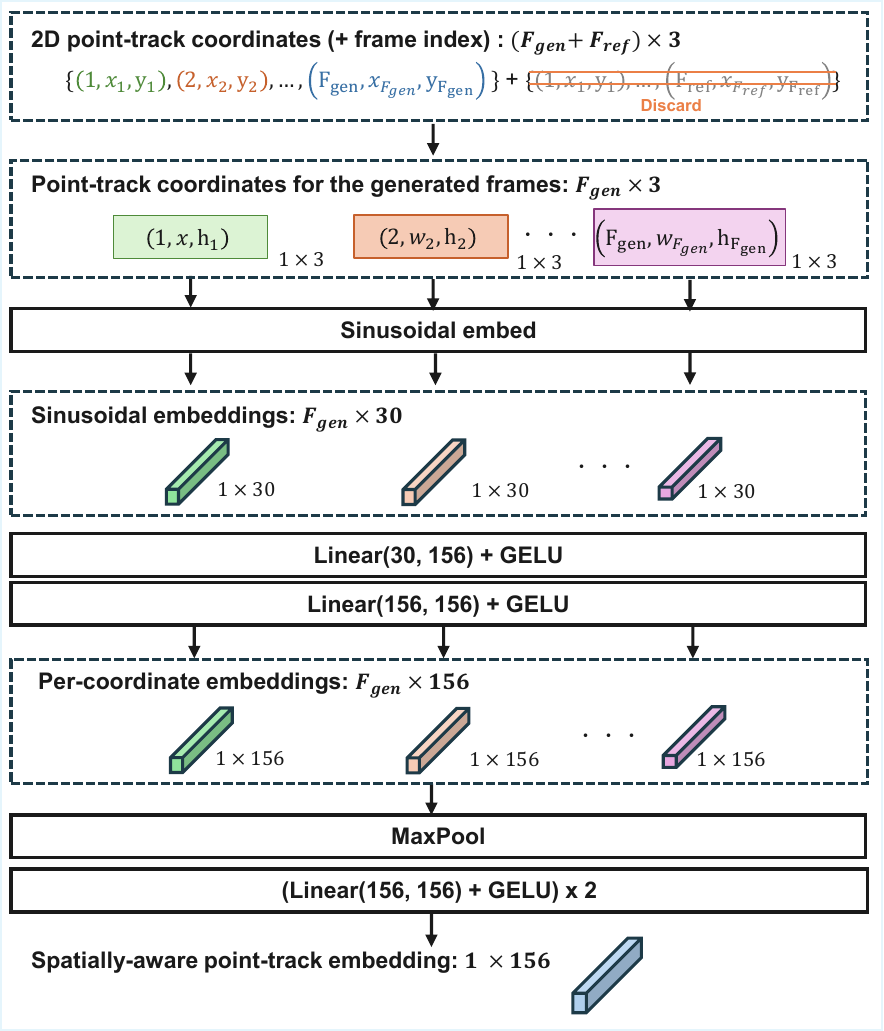} %
  \caption{\textbf{Details of point-track embedder.} For each point-track, we encode its 2D coordinates together with their corresponding frame indices using a sinusoidal positional embedding, followed by a shared coordinate-wise MLP to produce per-coordinate embeddings. These embeddings are then aggregated via temporal max pooling to obtain a single, unified, spatially-aware point-track embedding.}
  \label{fig:embedder_supp}
\end{figure}

\noindent \textbf{Spatially-aware point-track embedder.} A point-track is defined as a sequence of 2D coordinates spanning $F$ generated frames and $R$ reference images:
$\{(x^{\text{gen}}_i, y^{\text{gen}}_i)\}_{i=1}^{F}$ and $\{(x^{\text{ref}}_j, y^{\text{ref}}_j)\}_{j=1}^{R}$,
where $(x, y) \in [0, W] \times [0, H]$ denotes pixel coordinates.
Some coordinates may be undefined ($\varnothing$) (e.g., due to occlusion or filtering), which are excluded from the inputs, resulting in point-tracks of variable lengths.
Each valid coordinate is embedded as:
\begin{equation}
    \mathbf{p}_i = \text{MLP}\bigl(\text{SinusoidalEmb}(i, x^{\text{gen}}_i, y^{\text{gen}}_i)\bigr) \in \mathbb{R}^{156},
\end{equation}
where the sinusoidal embedding encodes both spatial location and frame index.
The per-coordinate embeddings are then aggregated via max pooling over all available coordinates to produce a unified per-track embedding:
\begin{equation}
    \mathbf{id} = \text{MaxPool}(\mathbf{p}_1, \ldots, \mathbf{p}_F) \in \mathbb{R}^{156}.
\end{equation}
Additional architectural details are provided in Fig.~\ref{fig:embedder_supp}.

\begin{figure}[!t] %
  \centering
  \includegraphics[width=\columnwidth]{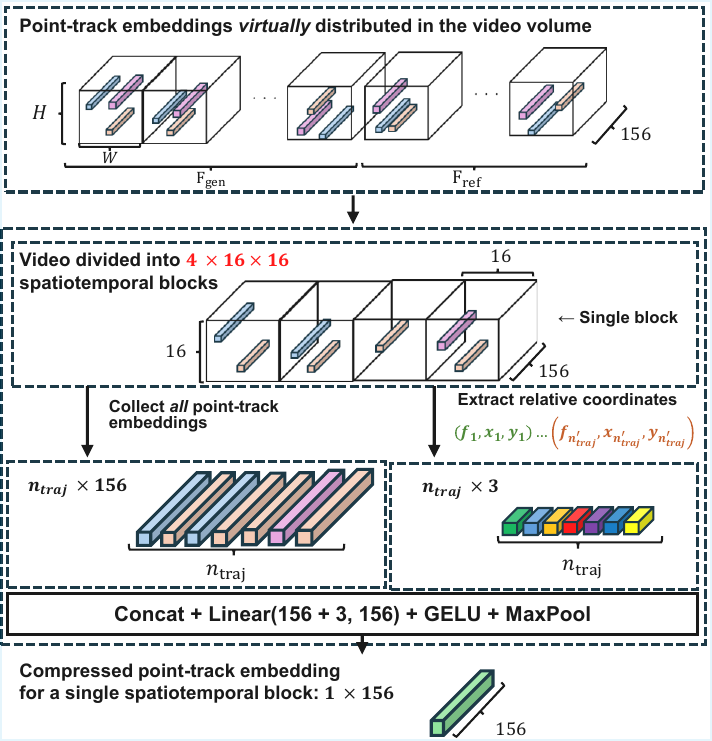}
  \caption{\textbf{Details of the point-track adapter.}
To align pixel-space point-track embeddings with the compressed patchified latent space (with $4 \times$ temporal and $16 \times 16$
spatial downsampling), our lightweight adapter partitions the video volume into non-overlapping $4 \times 16 \times 16$ spatiotemporal blocks. Within each block, point-track embeddings are concatenated with their relative intra-block coordinates and processed by an MLP, followed by max pooling to produce a single per-block latent-aligned token.}
  \label{fig:encoder_supp}
\end{figure}

\noindent \textbf{Point-track adapter.} To inject pixel-space point-track embeddings into the video diffusion model, we downsample them to match the compressed patchified latent space: $4\times$ temporally and $16\times16$ spatially.

\noindent We partition the pixel-space video volume ($F \times H \times W$) into non-overlapping $4 \times 16 \times 16$ spatiotemporal blocks.
Each block corresponds to a single patchified latent token at resolution $\frac{F}{4} \times \frac{H}{16} \times \frac{W}{16}$.
For each block, we gather the embeddings $\{\mathbf{id}_t\}_{t=1}^{n'}$ of all point-tracks whose coordinates fall within the block.
Inspired by the grid pooling mechanism in Point Transformer V2/V3~\cite{wu2022point, wu2024ptv3}, we aggregate these embeddings using max pooling.
However, naive max pooling discards precise positional information within each block.
To address this, we record the relative coordinate $(f'_t, x'_t, y'_t) \in [0, 4) \times [0, 16) \times [0, 16)$ of each point within its block, concatenate them with the corresponding track embedding, and apply an MLP prior to pooling:
\begin{equation}
    \mathbf{c} = \text{MaxPool}\bigl(\{\text{MLP}([\mathbf{id}_t; f'_t, x'_t, y'_t])\}_{t=1}^{n'}\bigr) \in \mathbb{R}^{156}.
\end{equation}
Our ablation study (see the numerical comparison between the columns \emph{Max pool only} and \emph{Max pool \& rel. pos.} in \cref{tab:ablation_traj}) confirms that incorporating relative positional encoding is crucial for precise motion controllability.

\noindent \paragraph{Computational efficiency.}
Our adapter operates on latent tokens and point-track embeddings instead of raw pixels, resulting in substantial computational savings.
For a video with $F=48$ frames at $480 \times 832$ resolution:
\begin{itemize}
    \item Latent tokens: $\frac{48}{4} \times \frac{480}{16} \times \frac{832}{16} = 18{,}720$
    \item Point-track coordinates: $48 \times 15{,}000 = 720{,}000$
    \item Full pixel count (avoided): $48 \times 480 \times 832 \approx 19.2\text{M}$
\end{itemize}
By operating directly in the compressed latent space and point-track embeddings, our adapter is significantly more efficient than pixel-space alternatives. In contrast, prior methods without a dedicated point-track adapter~\cite{wang2025, chu2025wan} simply downscale coordinates by a factor of 16 with rounding, which discards fine-grained spatial information and degrades motion controllability.

\noindent \paragraph{Reference frame handling.}
The first frame and each reference image are encoded independently without temporal compression.
To maintain compatibility with the $4×$ temporal pooling used in our adapter, each reference frame is replicated four times and treated as a static four-frame sequence. \\

\noindent Additional architectural details are provided in~\cref{fig:encoder_supp}.

\section{Experiment details}
\label{app:exps_results}

\noindent \textbf{Baselines.}
Below, we describe the evaluation settings used for each baseline in \cref{sec:experiments}:
\begin{itemize}
    \item Go-With-The-Flow~\cite{burgert2025}: We provide dense optical flow as input and apply zero noise degradation, ensuring that the model strictly follows the ground-truth motion of the video.

    \item DiffusionAsShader~\cite{gu2025diffusion}: To align with the original problem formulation of DiffusionAsShader, we exclude point tracks that are not visible in the first frame. We also experimented with modifying the input layers to incorporate point tracks that do not originate from the first frame; however, this modification did not yield performance improvements.

    \item ATI~\cite{wang2025}: Similar to DiffusionAsShader, we exclude point tracks that are not visible in the first frame. Consistent with findings from MotionStream~\cite{shin2025motionstream}, we observe that visual quality degrades as the number of input point tracks increases. Accordingly, we cap the number of point-tracks at 40.

    \item Tora~\cite{zhang2024tora}: We evaluate Tora only under the Task-3 setting, as it supports sparse point-track conditioning only. In practice, we randomly sample up to 10 point tracks, since performance deteriorates when a larger number of tracks is provided.

    \item Wan-Move~\cite{chu2025wan}: As Wan-Move adopts a similar methodology and architecture to ATI, we follow the same evaluation protocol used for ATI.
    
\end{itemize}

\noindent \textbf{Training time and resources.} We train our 14B 480p model for 7 days on a single node with 8×A100 GPUs. The 720p model is initialized from the 480p checkpoint and further trained for 7 days on a single node with 8×H200 GPUs. Training is parallelized using FSDP (Fully Sharded Data Parallel) to ensure efficient memory utilization and scalability.

\section{Quantitative results on TapVid3D-ADT}
\begin{table}[tb]
    \caption{\textbf{Comparisons to baselines on TapVid3D-ADT~\cite{tapvid3d}.} We compare against SOTA baselines. \textbf{Bold} indicates best results. The horizontal lines distinguish the different task settings. See~\cref{tab:main} for corresponding results on the DAVIS2017 dataset.}
    \label{tab:tapvid3d}
    \centering
    \scriptsize 
    \setlength{\tabcolsep}{1.2pt} 
    \renewcommand{\arraystretch}{1.15}
    \vspace{-5pt}
    \begin{tabular}{lccccccc}
        \toprule
        \multirow{2}{*}{Method} 
        & \multirow{2}{*}{Backbone}
        & \multicolumn{2}{c}{Visual fidelity} 
        & \multicolumn{3}{c}{Reconstruction accuracy} 
        & \multicolumn{1}{c}{Motion fidelity} \\
        
        \cmidrule(lr){3-4}\cmidrule(lr){5-7}\cmidrule(lr){8-8}
        & 
        & FID $\downarrow$ 
        & FVD $\downarrow$
        & LPIPS $\downarrow$ 
        & PSNR $\uparrow$ 
        & SSIM $\uparrow$
        & EPE $\downarrow$ \\

        \midrule
        \multicolumn{7}{l}{\textbf{Dense tracks}} \\
        
        ATI~\cite{wang2025} & \texttt{CogVideoX 5B} & 67.75 &   614.5 &     0.447 &     13.69 &     0.568 &      23.62\\
        
        DAS~\cite{gu2025diffusion} & \texttt{CogVideoX 5B} & 80.07 &   707.6 &     0.498 &     14.95 &     0.561 &      27.79\\
        
        GWTF~\cite{burgert2025} & \texttt{CogVideoX 5B} & 75.28 &   632.3 &     0.430 &     14.61  &     0.601 &      15.60\\

        Wan-Move~\cite{chu2025wan} & \texttt{Wan2.1 14B} & 61.37 & 533.2 &  0.427 &  15.91 &  0.593 & 11.18 \\
        
        \rowcolor{gray!10}
        \textbf{\method (ours)} & \texttt{Wan2.1 1.3B} & \underline{45.88} & 366.0 & 0.307 & \textbf{17.91} & 0.659 & 5.342\\
        
        \rowcolor{gray!10}
        \textbf{\method (ours)} & \texttt{Wan2.1 14B} & 46.56 & \underline{361.7} & \underline{0.297} & 17.17 & \textbf{0.676} & \textbf{4.282}\\
        
        \rowcolor{gray!10}
        \textbf{\method (ours)} & \texttt{Wan2.2 14B} & \textbf{41.00} & \textbf{314.3} & \textbf{0.291} & \underline{17.84} & \underline{0.673} & \underline{4.429}\\
        \bottomrule
    \end{tabular}
\vspace{-5pt}
\end{table}

In addition to the DAVIS 2017 evaluation presented in the main paper~\cref{tab:main}, we compare our method against baseline approaches on the more challenging TapVid3D-ADT dataset. As shown in~\cref{tab:tapvid3d}, our model consistently outperforms all baselines. Notably, the performance margin is larger than that observed on DAVIS 2017, further demonstrating the effectiveness and robustness of our approach.

\section{Additional ablation study}
\label{app:ablation}

\subsection{Classifier-free guidance}
To evaluate the impact of classifier-free guidance (CFG), we apply classifier-free guidance over text prompts and reference images during inference and vary the guidance scale using $\text{cfg} \in \{1.5, 3.0, 5.0, 7.0\}$. The results are presented in \cref{tab:cfg}.
We observe that a high guidance scale $\text{cfg}=5.0$ is overly aggressive and occasionally leads to image saturation, whereas a low guidance scale $\text{cfg}=1.5$ results in slightly reduced video quality. Based on this trade-off, we adopt $\text{cfg}=3.0$ for all experiments. Furthermore, we find that applying CFG solely to the text prompt yields suboptimal performance. Instead, jointly applying CFG to both reference images and text prompts produces more stable and robust results.

\begin{table}[!t]
    \caption{\textbf{CFG-guidance strategy} We observe that applying classifier-free guidance (CFG) solely to the text prompt yields suboptimal performance. In our experiments, we instead apply CFG jointly to both the reference images and text prompts, using a guidance scale of $3.0$. This strategy consistently produces stable and robust results across all evaluated datasets. }
    \centering
    \scriptsize
    \setlength{\tabcolsep}{2.5pt} %
    \begin{tabular}{lcccccc}
        \toprule
        & \multicolumn{2}{c}{{Visual fidelity}} 
          & \multicolumn{3}{c}{{Reconstruction accuracy}} 
          & \multicolumn{1}{c}{{Motion fidelity}} \\
        \cmidrule(r){2-3}\cmidrule(lr){4-6}\cmidrule(l){7-7}
        Method 
          & FID $\downarrow$ & FVD $\downarrow$
          & LPIPS $\downarrow$ & PSNR $\uparrow$ & SSIM $\uparrow$
          & EPE $\downarrow$ \\
        \midrule

        \multicolumn{7}{l}{\textbf{Joint CFG over reference and text prompts}} \\ %
        \hspace{3mm} cfg = 1.5
          & \textbf{27.60} & \textbf{306.9} & \textbf{0.259} & \underline{17.07} & \underline{0.580} & 7.342 \\
        \rowcolor{gray!10}        
        \hspace{3mm} cfg = 3.0
          &  \underline{28.03}  & 333.1 & 0.268 & 16.74 & 0.579 & \textbf{7.112} \\
        \hspace{3mm} cfg = 5.0 
          & 28.85 & 374.5 & 0.285 & 15.94 & 0.562 & \underline{7.140}  \\
        \hspace{3mm} cfg = 7.0
          & 33.91 & 486.8 & 0.348 & 13.56 & 0.511 & 7.345 \\

        \multicolumn{7}{l}{\textbf{CFG over text prompts only}} \\ %
        \hspace{3mm} cfg = 1.5
          & 29.85 & 318.5 & 0.269 & 16.94 & 0.567 & 7.482 \\
        \hspace{3mm} cfg = 3.0
          &  30.19  & 348.1 & 0.273 & 16.57 & 0.561 & 7.604 \\
        \hspace{3mm} cfg = 5.0 
          & 33.37 & 420.7 & 0.289 & 15.81 & 0.545 & 7.683  \\
        \hspace{3mm} cfg = 7.0
          & 36.85 & 469.5 & 0.309 & 15.13 & 0.527 & 7.728 \\

        \multicolumn{7}{l}{\textbf{CFG over reference image only}} \\ %
        \hspace{3mm} cfg = 1.5
          & 28.31 & \underline{313.4} & \underline{0.261} & \textbf{17.11} & \textbf{0.582} & 7.195 \\
        \hspace{3mm} cfg = 3.0
          &  28.57  & 337.1 & 0.268 & 16.95 & \textbf{0.582} & 7.310 \\
        \hspace{3mm} cfg = 5.0 
          & 30.05 & 348.9 & 0.279 & 16.71 & 0.570 & 7.736  \\
        \hspace{3mm} cfg = 7.0
          & 32.16 & 381.6 & 0.296 & 16.29 & 0.553 & 8.098 \\
        \bottomrule
    \end{tabular}
    \label{tab:cfg}
\end{table}

\subsection{Qualitative ablation analysis}
\label{sec:ablation_qual_supp}
\cref{fig:ablation_sup} presents qualitative results corresponding to the ablations in \cref{tab:ablation_traj}. Although the task is simply to generate a zoomed-in view of the TV given a reference image of the zoomed-out conference room, we observe that the video models struggle to perform this task without our proposed design choices.

\noindent In particular, when using random point-track embeddings, the model fails to follow the given point-tracks, resulting in a substantially different zoom-in behavior. 
Without the point-track adapter (\textit{Random embeddings} in \cref{tab:ablation_traj}) or relative position injection (\textit{Max pool only} in \cref{tab:ablation_traj}), the model still struggles to adhere precisely to the provided tracks, producing a tilted TV due to the loss of fine-grained motion details.

\noindent We further find that training on datasets with ground-truth annotations is crucial. When trained solely on real video datasets with noisy point-track labels, the model catastrophically fails to generate the desired zoomed-in video. We hypothesize that this occurs because models trained exclusively on noisy annotations do not learn to strictly follow motion cues. Instead, they implicitly treat the input tracks as unreliable and attempt to “correct” them, leading to undesirable overfitting to specific patterns rather than faithfully following the provided point-track conditions.

\begin{figure*}[!t]
    \centering
        \captionsetup{type=figure}
        \includegraphics[width=\textwidth]{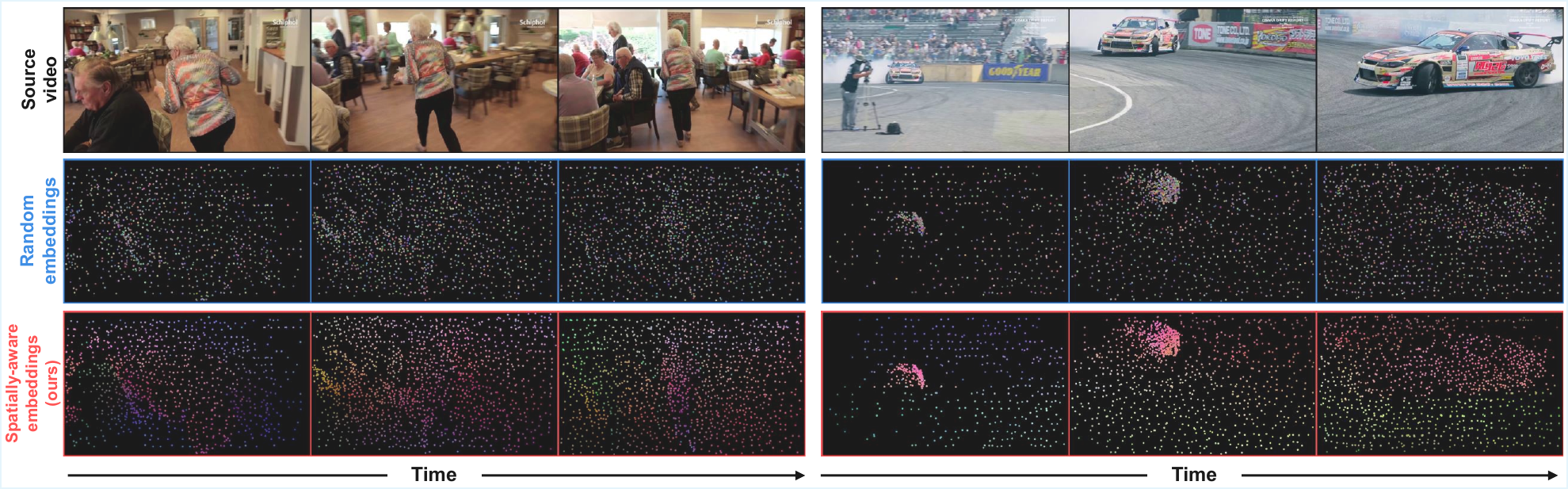}
        \captionof{figure}{\textbf{PCA visualization of point-track embeddings.} We visualize both random embeddings and our spatially-aware point-track embeddings using PCA and project the resulting components onto the pixel space. Our embeddings exhibit clear spatial correlations, whereas random embeddings show no meaningful spatial structure.}
    \label{fig:supp_pca}
\end{figure*}

\subsection{PCA analysis of point-track embeddings}
A key design choice in our framework is the use of spatially-aware point-track embeddings instead of random embeddings. This design yields a substantial performance improvement, as shown in~\cref{tab:ablation_traj} (\textit{Random embeddings} vs.\ \textit{Spatial-aware embeddings}). To further analyze this effect, we perform a PCA visualization of both random embeddings and our proposed embeddings. As shown in \cref{fig:supp_pca}, our embeddings exhibit clear spatial correlations, whereas random embeddings do not preserve meaningful spatial structure. These results suggest that incorporating spatially correlated point-track embeddings is crucial for effectively incorporating reference-anchored point-track conditioning into our video model.

\section{Details of Qualitative Comparison}
\label{qual_detail_supp}
In this section, we present additional qualitative results. We encourage readers to view the accompanying videos on our supplementary webpage, which contains more examples than can be included in the paper.

\begin{figure*}[!t]
    \centering
        \captionsetup{type=figure}
        \includegraphics[width=\textwidth]{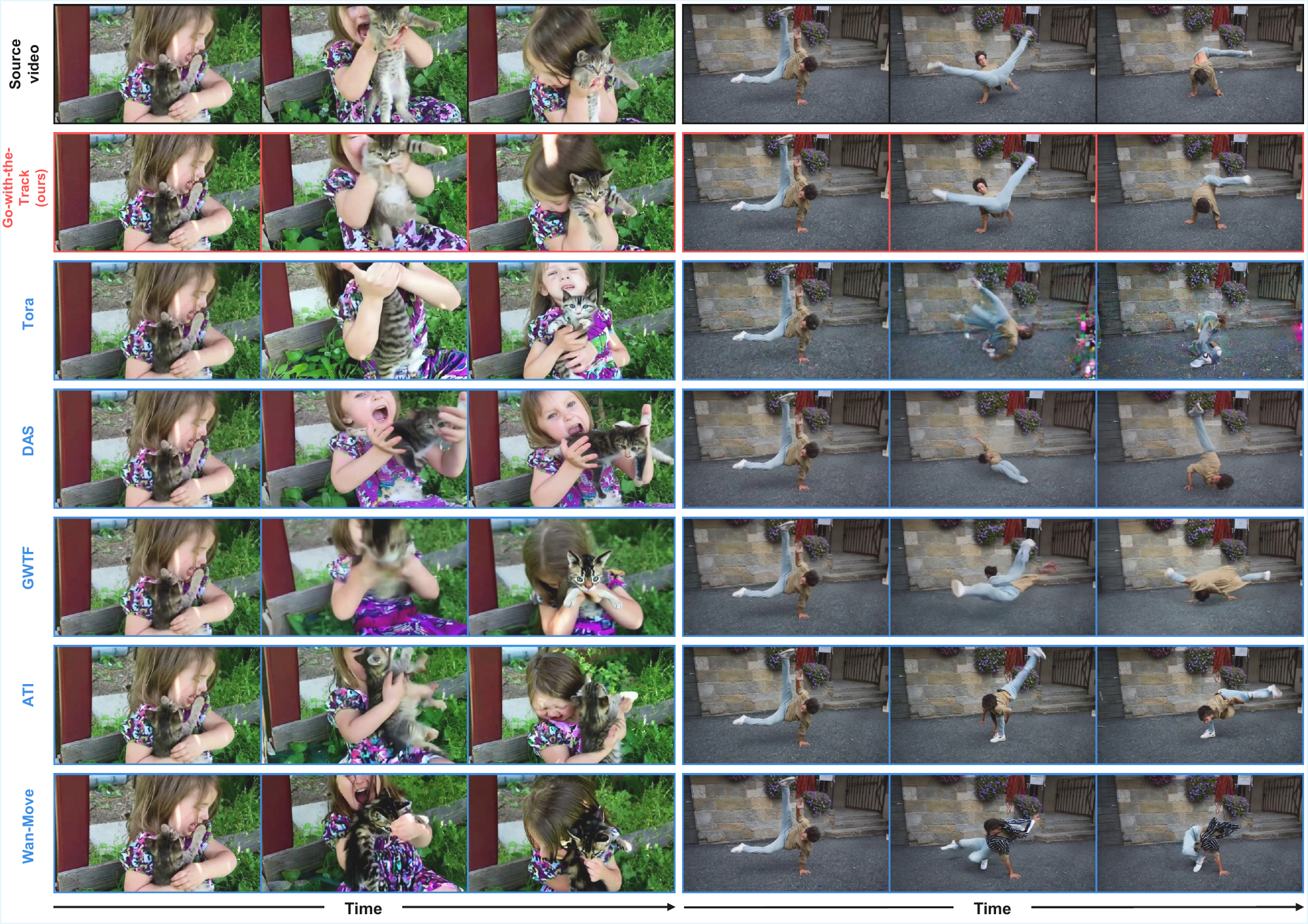}
        \captionof{figure}{\textbf{Qualitative comparisons to baselines on video reconstruction.}  Given the first frame and extracted point tracks, each method attempts to reconstruct the source video.
        As evident, we find the outputs from \method align much more closely with ground-truth while preserving the spatial structure and element identity better.
        Additional results are provided in the supplementary webpage.}
    \label{fig:supp_davis_recon}
\end{figure*}

\begin{figure*}[!t]
    \centering
        \captionsetup{type=figure}
        \includegraphics[width=\textwidth]{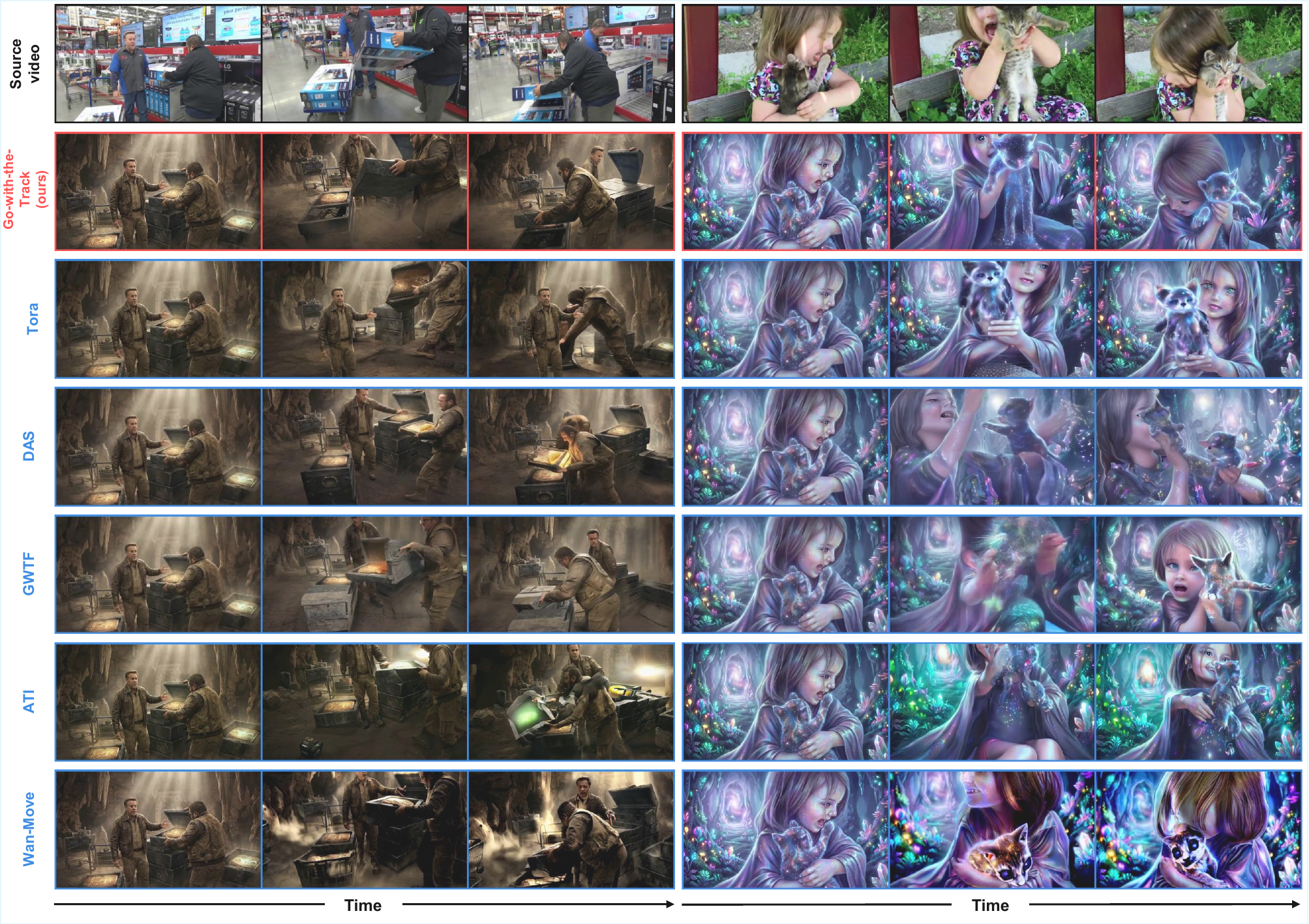}
        \captionof{figure}{\textbf{Qualitative comparisons to baselines on video restylization.} Given stylized first frames and point-track conditions, each method generates motion-preserved restylized videos. \method better preserves the source motion while adhering to the appearance of the stylized first frame. Additional results are provided in the supplementary webpage.}
    \label{fig:supp_davis_stylize}
\end{figure*}

\noindent \textbf{Baseline comparison.} We provide visual comparisons between all baselines and our \method under the conventional point-track-conditioned first-frame-to-video generation setting. Specifically, we present two types of comparisons:

\begin{enumerate}[label={\alph*)}]
    \item \textbf{Reconstruction:} Given a source video, we extract point-track conditions using an off-the-shelf point tracker~\cite{ngo2024delta} and use the first frame as the reference image, aiming to reconstruct the original video. As shown in \cref{fig:supp_davis_recon}, \method more closely aligns with the ground truth, better preserving spatial structure and object identity.
    \item \textbf{Restylization:} Given a source video, we similarly extract point-track conditions but stylize the first frame before providing it to the model. As illustrated in \cref{fig:supp_davis_stylize}, our model faithfully preserves the motion of the source video while adhering to the appearance of the stylized first frame.
\end{enumerate}

\noindent Additional visual demonstrations are provided in our supplementary webpage.\\

\begin{figure*}[!t]
\centering
\captionsetup{type=figure}
\includegraphics[width=\textwidth]{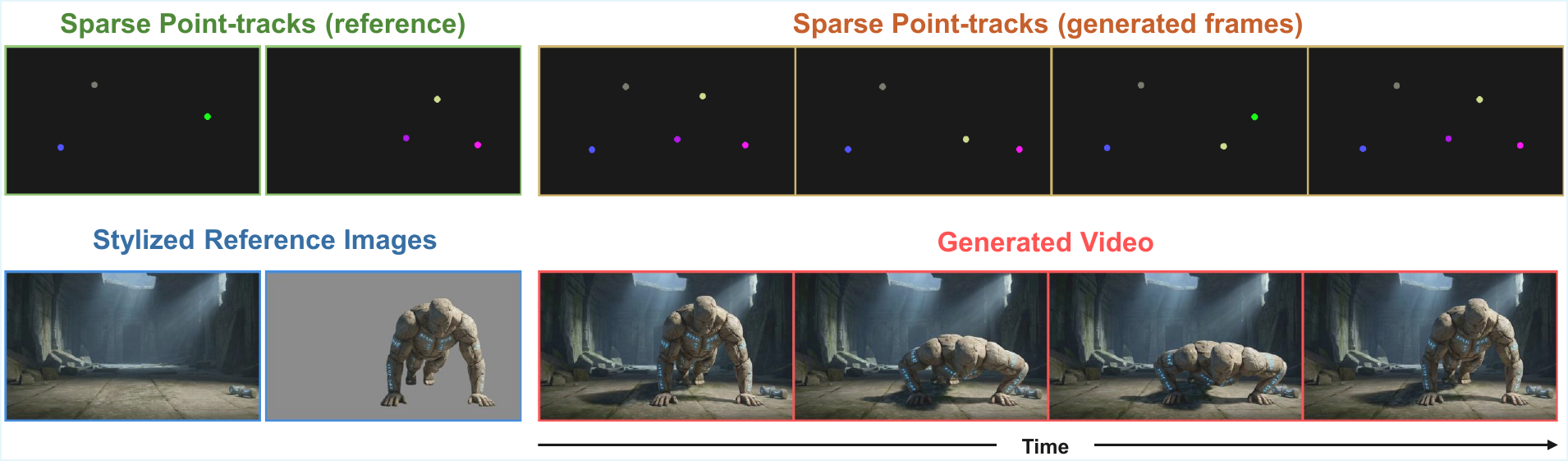}
\captionof{figure}{\textbf{Examples with Sparse Point-tracks.} Although the model is not explicitly trained on extremely sparse point-track inputs, our model successfully generates videos that adhere to the given motion conditions.}
\label{fig:sparse_res}
\end{figure*}

\noindent \textbf{Sparse Point-tracks.} We present visual results under extremely sparse point-track conditions in Fig.~\ref{fig:sparse_res}. Although our model is not explicitly trained with such a small number of point tracks, it is still able to generate videos that faithfully follow the motion conditions. This observation is consistent with the findings of~\citep{geng2025motionprompting}.

\begin{figure*}[!t] 
  \centering
  \includegraphics[width=\textwidth]{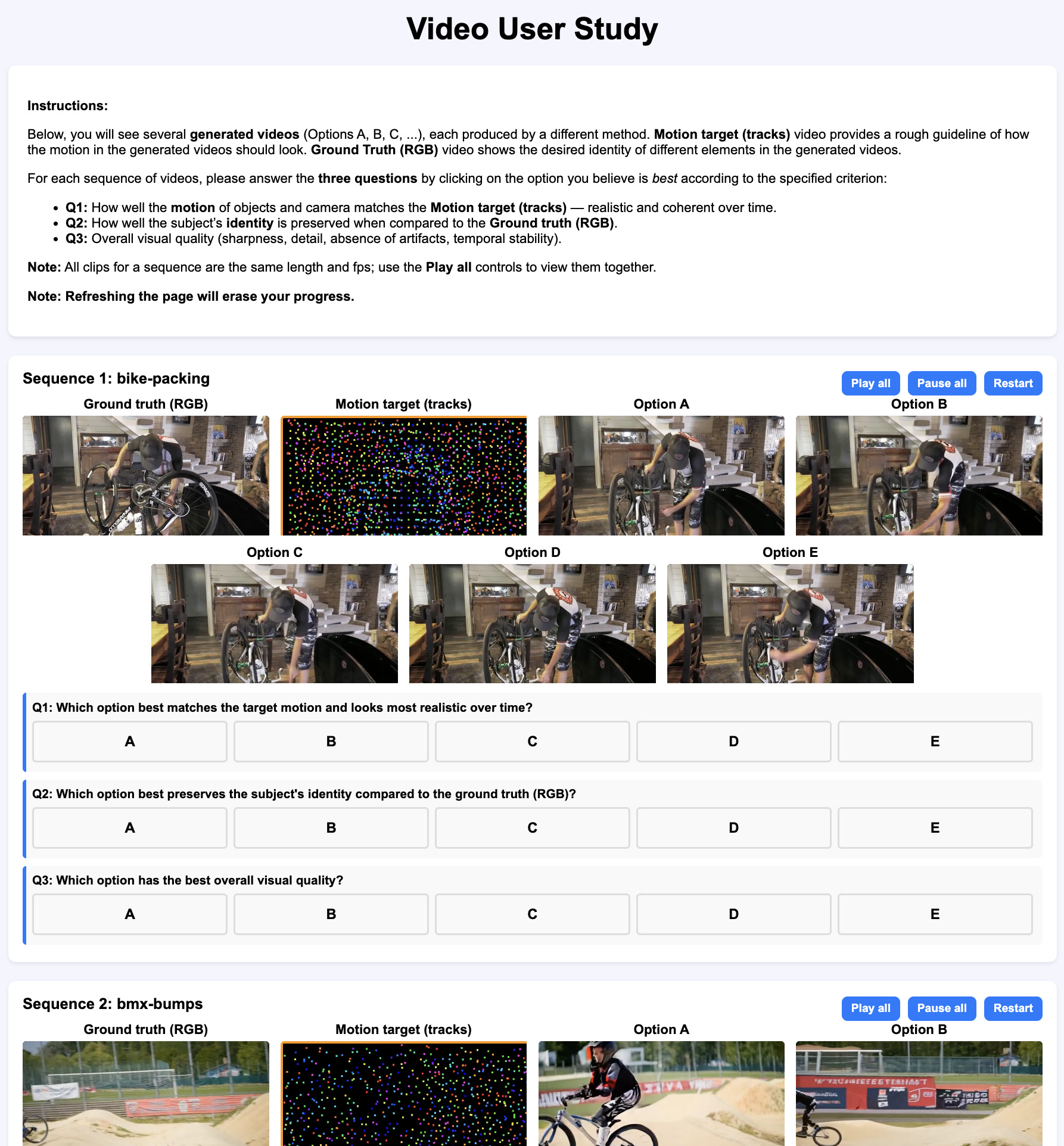}
  \caption{\textbf{User Study.} At the top, we show participants instructions on how to perform the task and answer the questions. Then, we provide an interactive interface that allows synchronous playback of all videos side by side while answering three questions about motion following, subject identity preservation, and overall quality for each example. Each participant annotates 30 examples by answering 90 questions in total.}
  \label{fig:user_study}
\end{figure*}

\noindent \textbf{User study.} We conducted an anonymous user study with 45 participants to qualitatively compare \method against baseline methods. Each participant evaluated 30 samples. For each sample, participants were shown the input point tracks, the ground-truth video, and randomly shuffled outputs from the baselines and \method. They were asked to rate each result based on three criteria: (a) motion adherence, (b) subject identity preservation, and (c) overall visual quality. A snapshot of the survey interface is shown in Fig.~\ref{fig:user_study}.

\section{Details of Applications}
\label{sec:sup_app_detail}

\noindent Please refer to our supplementary webpage for more visuals of each application.

\begin{figure*}[t] %
  \centering
  \vspace{-10pt}
  \includegraphics[width=0.91\textwidth]{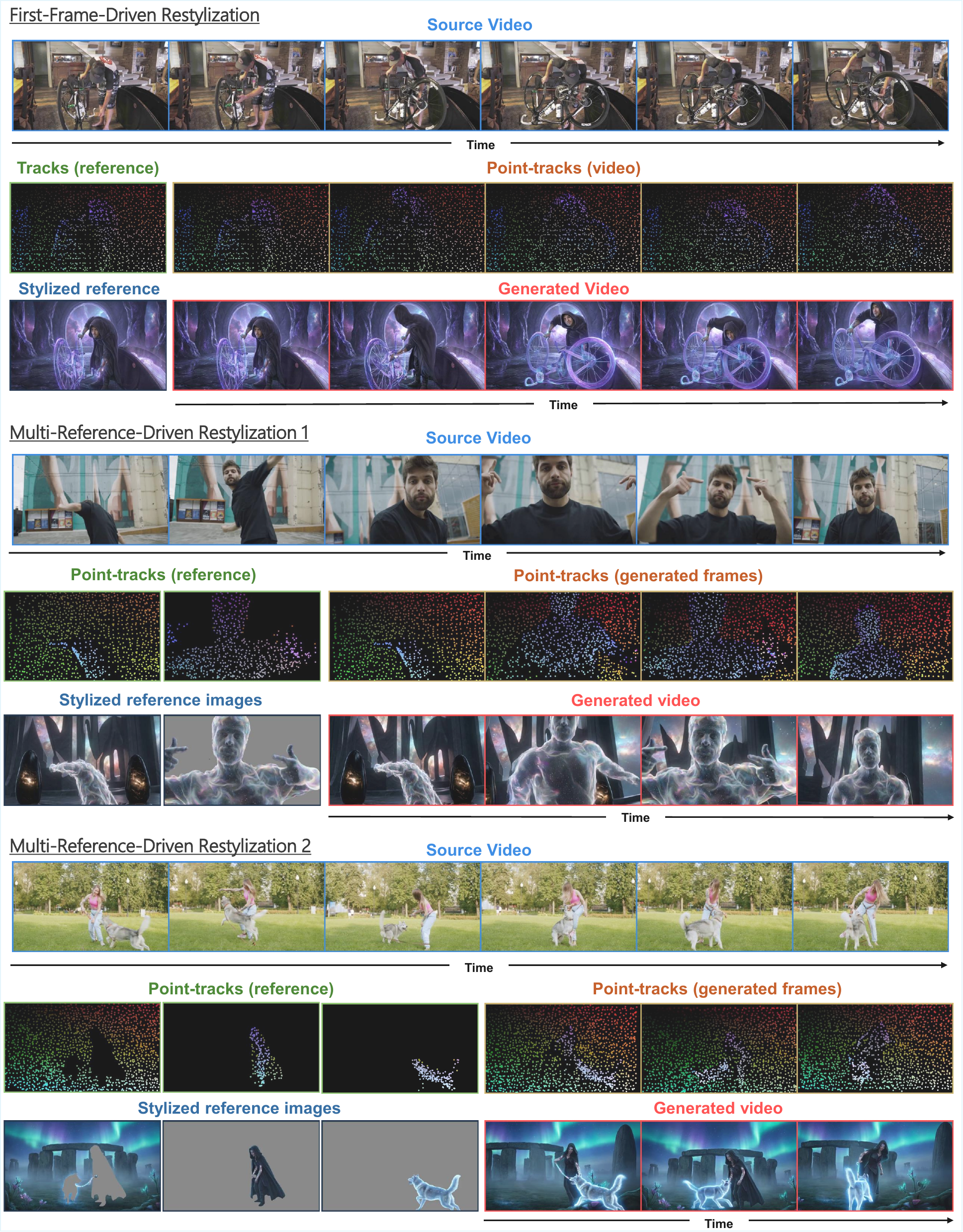} %
  \vspace{-10pt}
  \caption{\textbf{Video restylization.} Given point-tracks estimated from a source video, along with stylized reference images, \method produces restylized videos while preserving the original motion. Additional results are available on our supplementary website.}
  \label{fig:supp_video_restylize}
  \vspace{-10pt}
\end{figure*}

\noindent \textbf{Video restylization.} Given a source video, we first estimate point tracks using an off-the-shelf point tracker~\cite{ngo2024delta}, following the procedure described in \cref{app:iterative}. We then generate edited reference images from selected frames of the source video using NanoBanana~\cite{gemini25flashimage2025}. 
The reference images are either full frames from the source video or cropped objects extracted from it. For the latter, we employ SAM 3~\cite{carion2025sam3segmentconcepts} to segment and crop the target object, and gray out pixels outside the selected region.
Using these edited images as references, together with the estimated point tracks, \method generates a variety of restylized videos while preserving the underlying motion. Additional visual results are available on our supplementary webpage and \cref{fig:supp_video_restylize}.

\begin{figure*}[t] %
  \centering
  \vspace{-10pt}
  \includegraphics[width=0.95\textwidth]{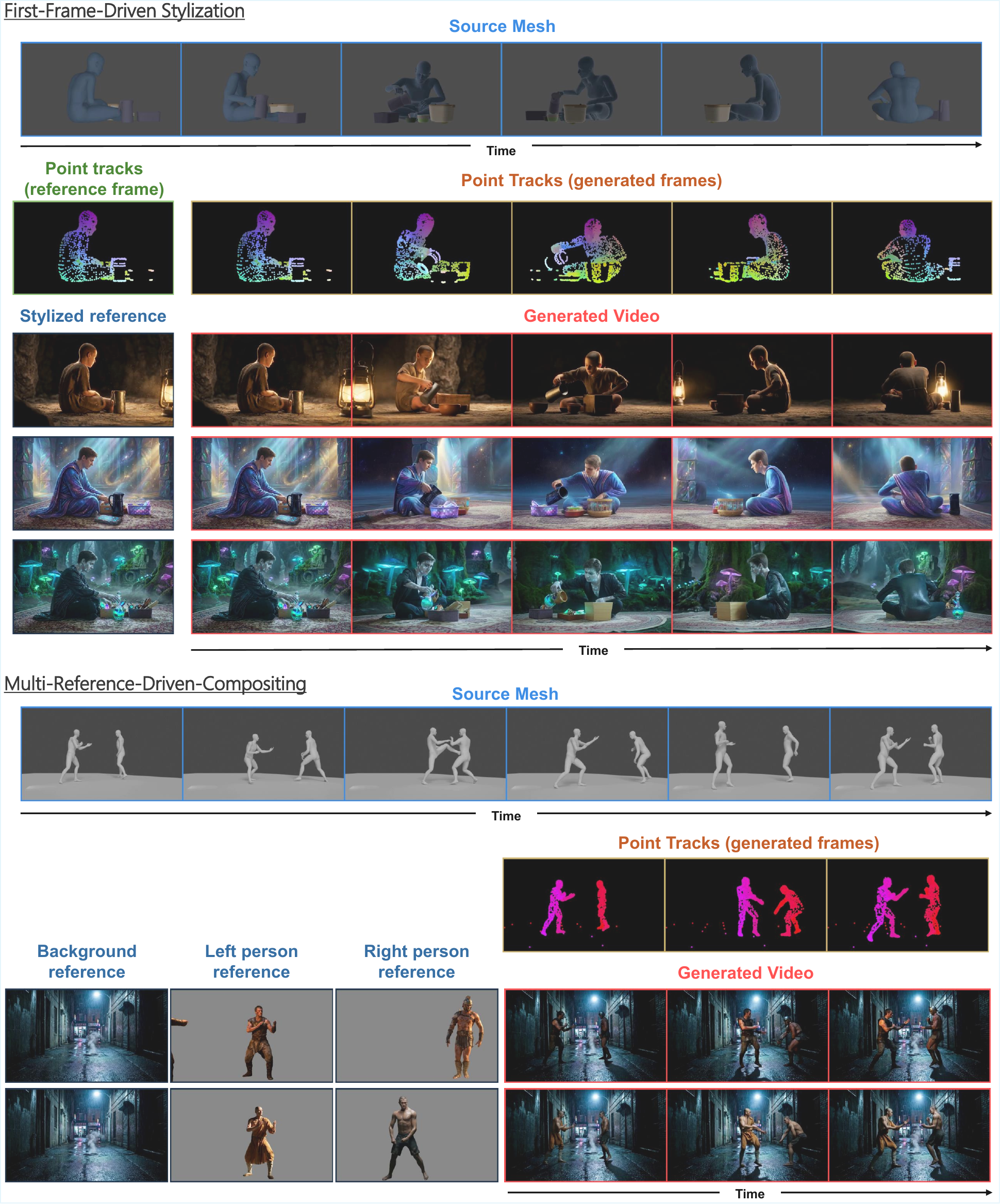} %
  \vspace{-10pt}
  \caption{\textbf{Mesh-driven compositing and stylization.} We render static and dynamic mesh scenes from arbitrary viewpoints and stylize the rendered images to obtain reference images. Together with point tracks derived from projected mesh vertices, \method generates stylized mesh-animated videos. Additional results are shown in our supplementary webpage.}
  \vspace{-10pt}
  \label{fig:supp_video_mesh}
\end{figure*}

\noindent \textbf{Mesh-driven compositing and stylization.}
We render textureless static and dynamic mesh scenes from arbitrary viewpoints using Blender. The rendered images are then stylized with NanoBanana~\cite{gemini25flashimage2025} to produce diverse visual appearances. To obtain point-tracks, we project mesh vertices onto the image plane along a predefined camera trajectory and use their resulting 2D trajectories as point tracks. Conditioned on these point-tracks and the stylized reference images, \method generates stylized mesh-animated videos. Additional visual results are provided on our supplementary webpage and \cref{fig:supp_video_mesh}.

\begin{figure*}[t] %
  \centering
  \vspace{-10pt}
  \includegraphics[width=0.8\textwidth]{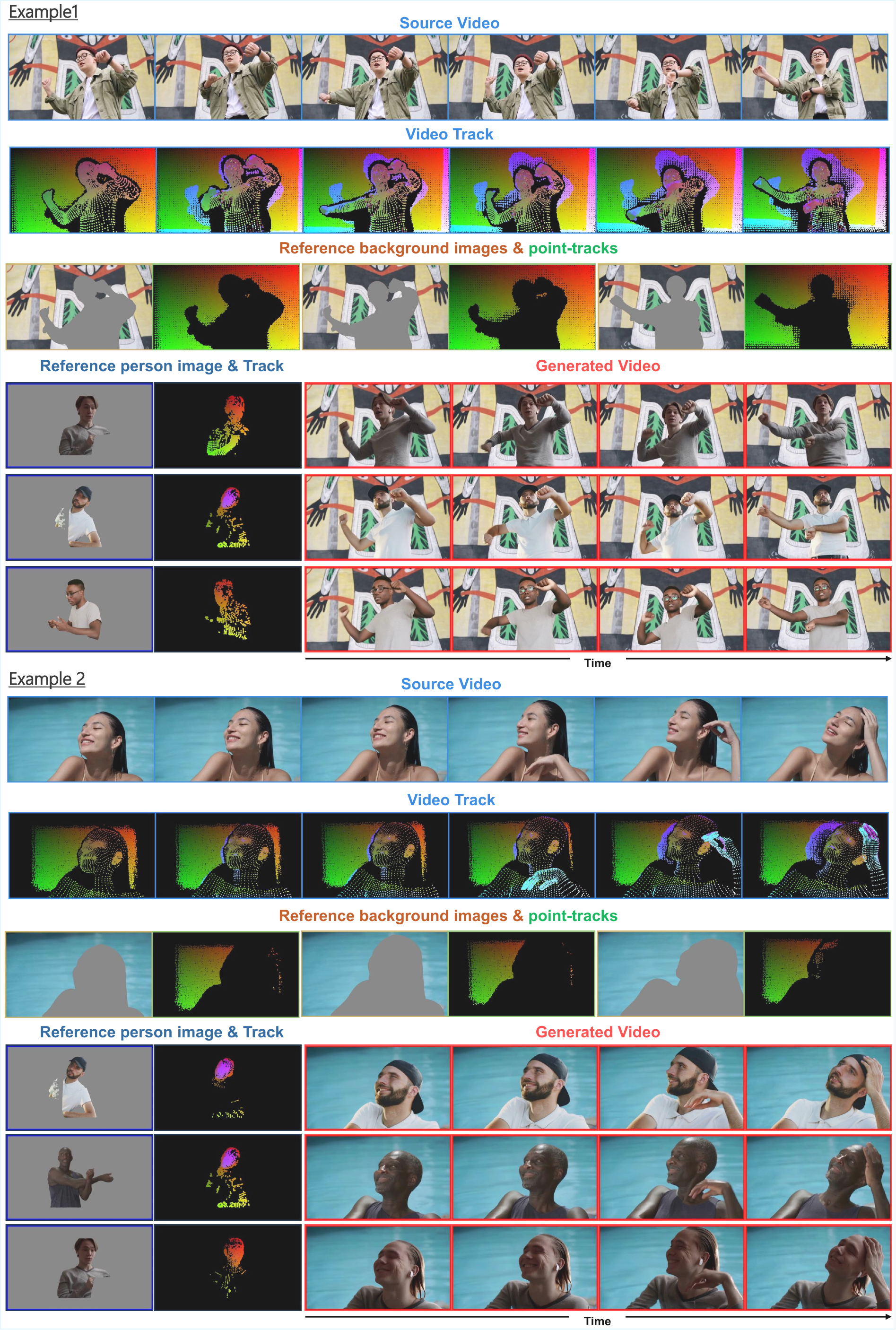} %
  \vspace{-10pt}
  \caption{\textbf{Keypoint-driven compositing.} Given a source video and a reference image, facial and full-body keypoints are extracted to form reference-anchored point tracks. Conditioned on these keypoint-derived point-tracks and the reference image, \method transfers the reference subject’s appearance to the source video while preserving the original motion. Additional results are available on our supplementary webpage.}
  \label{fig:supp_video_keypoint}
  \vspace{-50pt}
\end{figure*}

\noindent\textbf{Keypoint-driven compositing.} Given a human-centric video and a reference image, we apply off-the-shelf facial and full-body keypoint detectors~\cite{giebenhain2025pixel3dmm, yang2026sam3dbody} to extract anatomical keypoints across video frames and the reference image. Since keypoints sharing the same index correspond to the same semantic landmark (e.g., the left eye) across different images, they naturally form our reference-anchored point tracks.
Conditioned on these keypoint-derived point-tracks and the reference image, \method transfers the appearance of the reference subject onto the source video while faithfully preserving the original motion dynamics. Additional visual results are provided on our supplementary webpage and \cref{fig:supp_video_keypoint}.

\begin{figure*}[t] %
  \centering
  \includegraphics[width=0.82\textwidth]{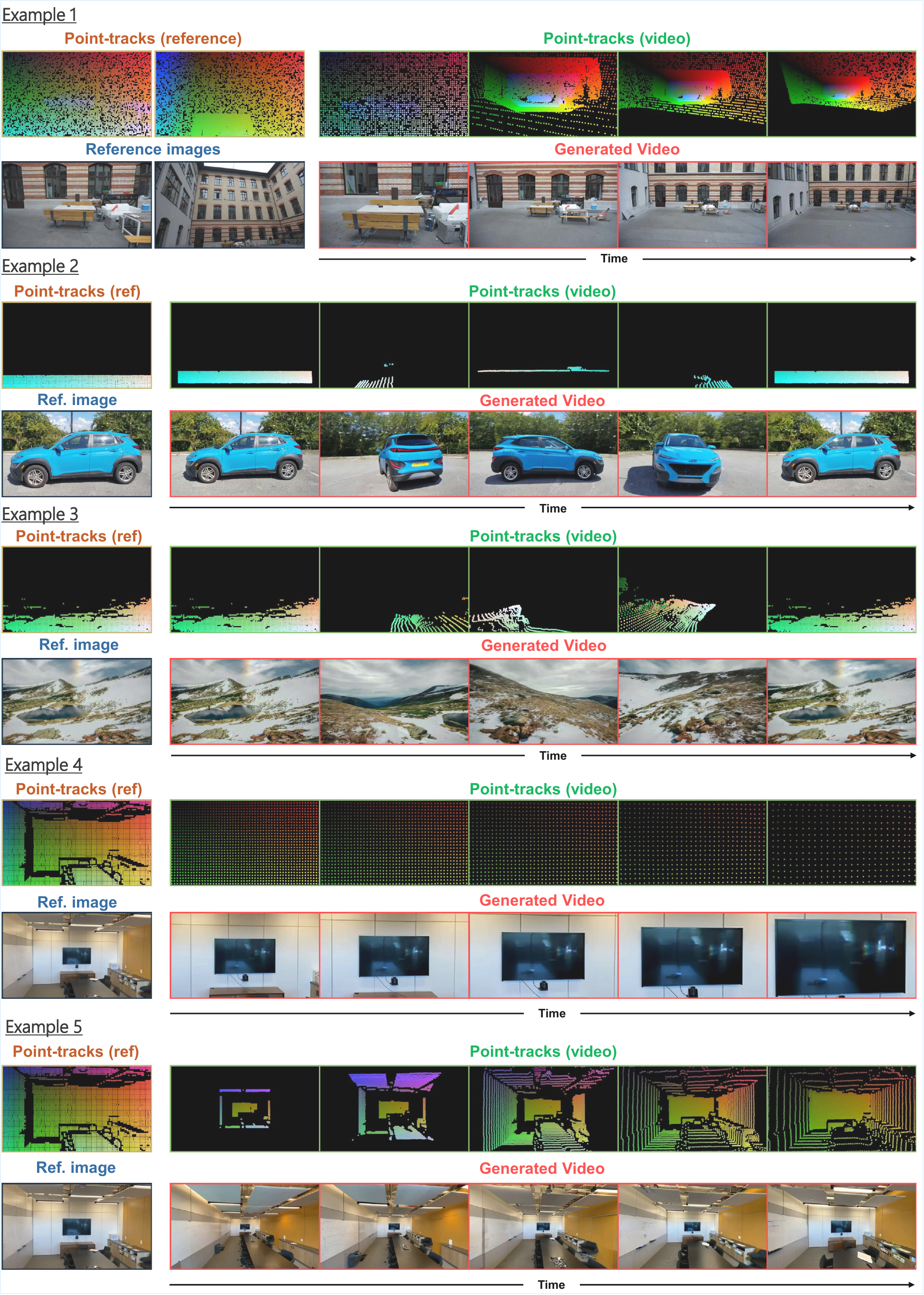} %
  \caption{\textbf{Camera control in static  scenes.} Reprojected 3D point clouds, together with reference images captured from arbitrary viewpoints, enable \method to retarget camera motion in both static scenes along user-defined trajectories. Additional results are available on our supplementary webpage.}
  \label{fig:supp_video_camera_static}
  \vspace{-50pt}
\end{figure*}

\begin{figure*}[t] %
  \centering
  \includegraphics[width=0.86\textwidth]{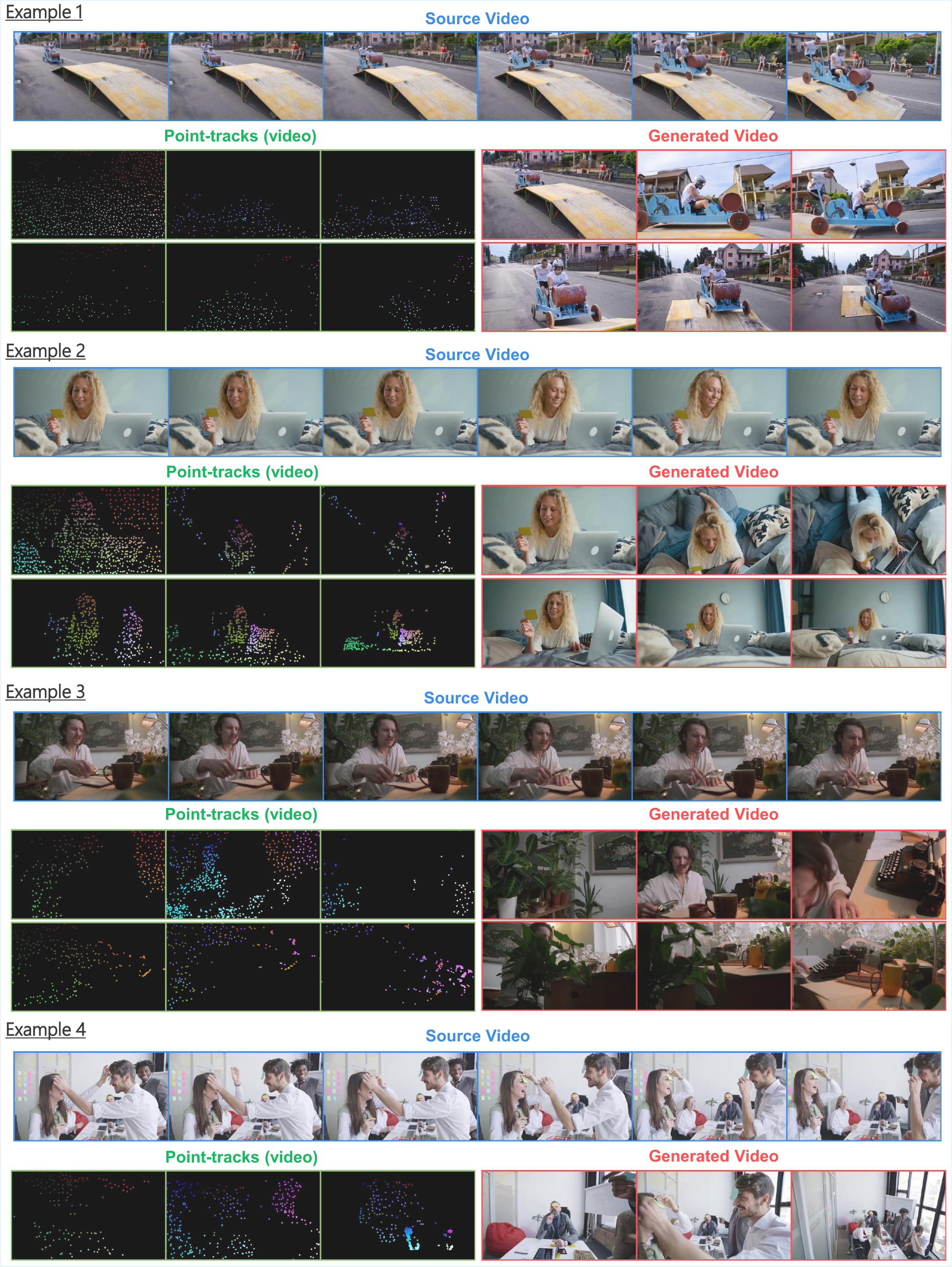} %
  \caption{\textbf{Camera control in dynamic scenes.} Reprojected dynamic point clouds, together with reference images captured from arbitrary viewpoints, enable \method to retarget camera motion in dynamic scenes along user-defined trajectories. Additional results are available on our supplementary webpage.}
  \label{fig:supp_video_camera_dynamic}
  \vspace{-50pt}
\end{figure*}

\noindent \textbf{Camera control in static and dynamic scenes.} In addition to the input video of the scene, we use $\pi^3$~\cite{wang2025pi} or DELTA~\cite{ngo2024delta} based point-tracks while rendering it from a new camera trajectory.

\noindent\textbf{Camera control in static and dynamic scenes.}
\method enables camera retargeting for both static and dynamic scene videos, given multiple reference images captured from different viewpoints.

\noindent For static scenes, we first reconstruct a 3D point cloud and estimate camera poses using $\pi^3$~\cite{wang2025pi}. The reconstructed point cloud is then reprojected onto user-defined target camera trajectories. We apply z-buffering to remove occluded points in each rendered frame.

\noindent For dynamic scenes, we perform 3D tracking using DELTA~\cite{ngo2024delta} with iterative track densification (\cref{app:iterative}). We estimate per-frame camera parameters using $\pi^3$, and reproject the tracked points under custom camera trajectories.
To prevent noisy reprojections, we discard reprojected points from each frame that were marked as invisible in the original frame. 
We uniformly sample four frames from the source video as reference images for conditioning.

\noindent Additional visual results are provided on our supplementary webpage,  \cref{fig:supp_video_camera_static}, and \cref{fig:supp_video_camera_dynamic}.

\begin{figure*}[t] %
  \centering
  \includegraphics[width=\textwidth]{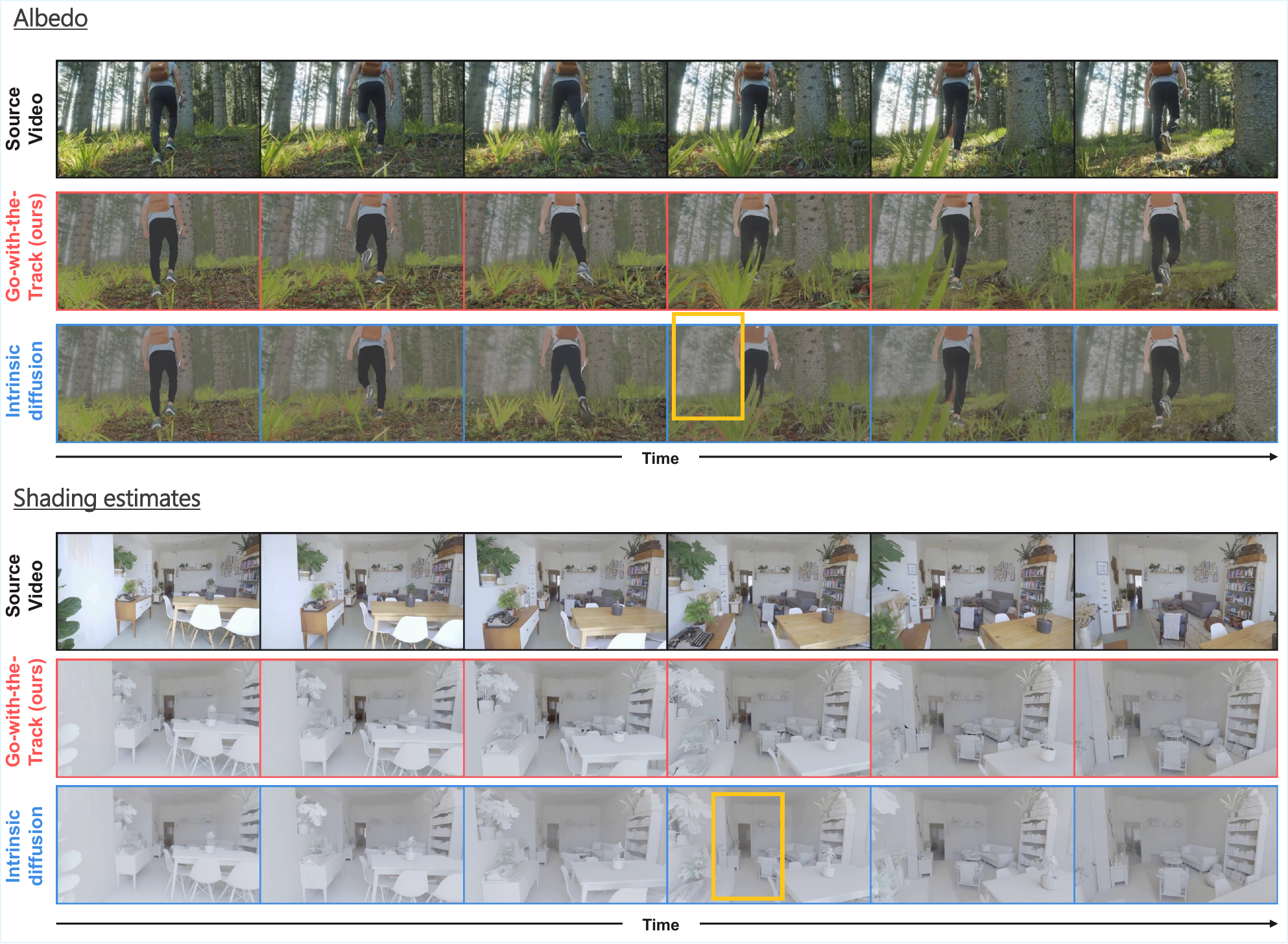} %
  \caption{\textbf{Temporal stabilization for intrinsic decomposition.} Using the albedo and shading estimates from the first and last frames as references, \method propagates these predictions across the sequence to produce temporally consistent albedo and shading videos,  reducing flicker compared to frame-by-frame estimation. Additional results are available on our supplementary webpage.}
  \label{fig:supp_video_temp}
  \vspace{-50pt}
\end{figure*}

\noindent \textbf{Temporal stabilization for albedo and relighting prediction.} Recently, powerful models~\cite{Luo2024IntrinsicDiffusion,kocsis2024intrinsic} have been developed for inverse image tasks such as intrinsic image decomposition. While these models perform well on a per-frame basis, their output often exhibits temporal flicker when applied frame by frame. We demonstrate that using our method we can obtain smooth, temporally stable generations when provided with only the first and last frame. We use a re-implementation of \cite{Luo2024IntrinsicDiffusion} to estimate both albedo and shading at the first and last keyframes of the input video. For each modality, we leverage \method to propagate these reference keyframes to the entire sequence. Our visuals on the supplementary webpage and \cref{fig:supp_video_temp} are more stable than the per-frame estimates, showing generalization to these unseen modalities.

\end{document}